\documentclass{bmvc2k}

\graphicspath{{generated-images/}{images/}}
\DeclareGraphicsExtensions{.pdf,.jpeg,.png}
\usepackage[table]{xcolor}
\usepackage{gensymb}
\usepackage{multirow}

\newcommand{\crd}[1]{\boldsymbol{c}_{#1}}
\newcommand{\patch}[1]{\boldsymbol{P}(#1)}
\newcommand{\corresp}{\Psi}
\newcommand{\corrpt}[1]{\boldsymbol{d}_{#1}}
\newcommand{\img}{I}
\newcommand{\match}{(\crd{i}, \crd{i}')}
\newcommand{\prob}{p}
\newcommand{\lab}{l}
\newcommand{\loss}{L}

\newcommand{\refsec}[1]{Section~\ref{#1}}
\newcommand{\reffig}[1]{Figure~\ref{#1}}
\newcommand{\reftab}[1]{Table~\ref{#1}}
\newcommand{\qmarks}[1]{``#1"}

\newcommand{\tlup}{\cellcolor{green!25} $\uparrow$}
\newcommand{\tldown}{\cellcolor{red!25} $\downarrow$}

\definecolor{somegray}{rgb}{0.5, 0.5, 0.5}
\newcommand{\darkgrayed}[1]{\textcolor{somegray}{#1}}
\makeatletter
\newcommand*\titleheader[1]{\gdef\@titleheader{#1}}
\AtBeginDocument{%
  \let\st@red@title\@title
  \def\@title{%
    \bgroup\normalfont\large\centering\@titleheader\par\egroup
    \vskip.5em\st@red@title}
}
\makeatother

\titleheader{\darkgrayed{This paper has been accepted for publication at the\\
British Machine Vision Conference (BMVC), Cardiff, 2019.}}

\title{Matching Features without Descriptors: \\Implicitly Matched Interest Points}

\addauthor{Titus Cieslewski}{titus@ifi.uzh.ch}{1}
\addauthor{Michael Bloesch}{m.bloesch@imperial.ac.uk}{2}
\addauthor{Davide Scaramuzza}{sdavide@ifi.uzh.ch}{1}

\addinstitution{
 Dept. of Informatics and Neuroinformatics\\
 University of Zurich and ETH Zurich \\ 
 Switzerland
}
\addinstitution{
 Dept. of Computing\\
 Imperial College London\\
 United Kingdom
}

\runninghead{Cieslewski et al}{Implicitly Matched Interest Points}

\begin{document}

\maketitle

\begin{abstract}
The extraction and matching of interest points is a prerequisite for many geometric computer vision problems.
Traditionally, matching has been achieved by assigning descriptors to interest points and matching points that have similar descriptors.
In this paper, we propose a method by which interest points are instead already implicitly matched at detection time.
With this, descriptors do not need to be calculated, stored, communicated, or matched any more.
This is achieved by a convolutional neural network with multiple output channels and can be thought of as a collection of a variety of detectors, each specialised to specific visual features.
This paper describes how to design and train such a network in a way that results in successful relative pose estimation performance despite the limitation on interest point count.
While the overall matching score is slightly lower than with traditional methods, the approach is descriptor free and thus enables localization systems with a significantly smaller memory footprint and multi-agent localization systems with lower bandwidth requirements.
The network also outputs the confidence for a specific interest point resulting in a valid match.
We evaluate performance relative to state-of-the-art alternatives.
\end{abstract}

\begin{figure}
  \centering
  \includegraphics[width=\linewidth]{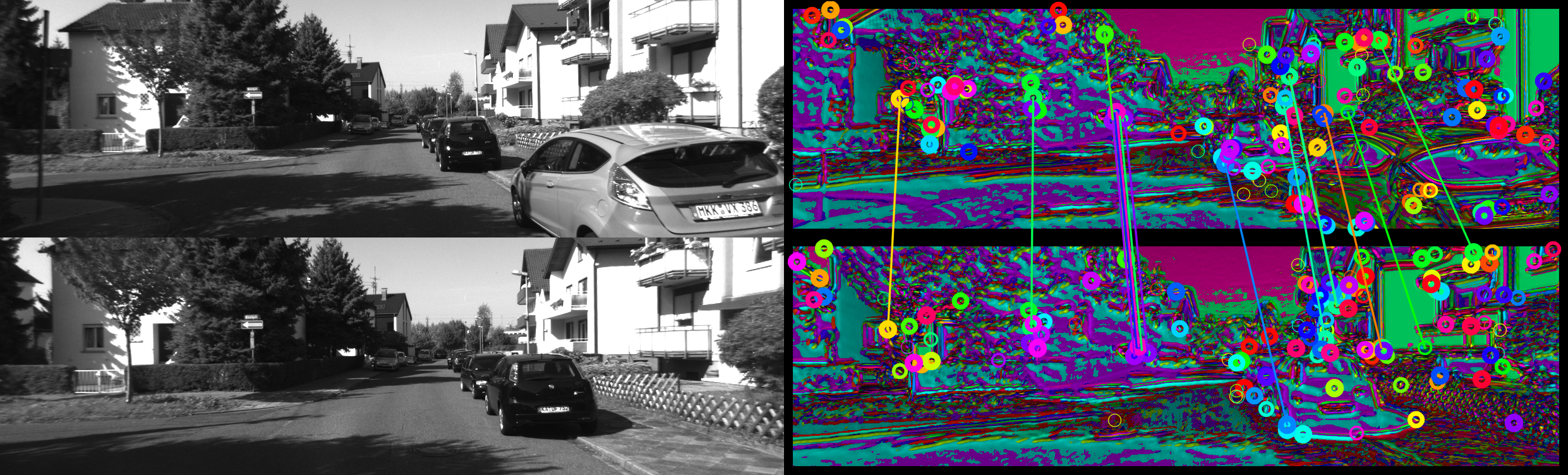}
  \vspace{-7mm}
  \caption{We propose a CNN interest point detector which provides implicitly matched interest points: descriptors are not needed for matching.
  This image illustrates the output of the network.
  Hue indicates which channel has the strongest response for a given pixel, and brightness indicates that response.
  Circles indicate the interest points, which are the global maxima of each channel.
  Lines indicate inlier matches after P3P localization.
  }
  \vspace{-5mm}
  \label{fig:eyecat}
\end{figure}

\section*{Multimedia Material}

Source code and data for this work are available at \\
\mbox{\url{https://github.com/uzh-rpg/imips_open}}.

\section{Introduction} 

Many applications of computer vision, such as structure from motion and visual localization, rely on the generation of point correspondences between images.
Correspondences can be found densely \cite{Horn81ai, Farnebaeck03scia, Rocco18nips}, where a correspondence is sought for every pixel, or with sparse feature matching, where correspondences are only established for a few distinctive points in the images.
While dense correspondences capture more information, it is often of interest to establish them only sparsely.
Sparse correspondences make algorithms like visual odometry or bundle adjustment far more tractable, both in terms of computation and memory.

Sparse feature matching used to be solved with hand-crafted descriptors \cite{Lowe04ijcv, Bay08cviu, Rublee11iccv}, but has more recently been solved using learned descriptors \cite{Yi16eccv, Detone18cvprw, Ono18nips}.
In this paper, we propose a novel approach that exploits convolutional neural networks (CNNs) in a new way.
Traditionally, features are matched by first detecting a set of interest points, then combining these points with descriptors that locally describe their surroundings, to form visual features.
Subsequently, correspondences between images are formed by matching the features with the most similar descriptors.
This approach, which has been developed with hand-crafted methods, has been directly adopted in newer methods involving CNNs.
As a result, different CNNs have been used for the different algorithms involved in this pipeline, such as interest point detection, orientation estimation, and descriptor extraction.

We propose a method that uses only a single, convolution-only neural network that subsumes all of these algorithms.
This network can be thought of as an extended interest point detector, but instead of outputting a single channel that allows the selection of interest points using non-maximum suppression, it outputs multiple channels (see \reffig{fig:eyecat}).
For each channel only the global maximum is considered an interest point, and the network is trained in such a way that the points extracted by the same channel from different viewpoints are correspondences.
As with traditional feature matching, geometric verification is required to reject outlier correspondences.
The network can alternatively be thought of as a dense point descriptor, but instead of expressing descriptors along the channel axis of the output tensor, each channel represents the response to a function defined in the descriptor space.

An important benefit of our method is that descriptors do not need to be calculated, stored, communicated or matched any more.
Previously, the minimal representation of an observation for relative pose estimation consisted of point coordinates and their associated descriptors.
With our method, the same observation can be represented with as little as the point coordinates ($3$ bytes for up to $4096 \times 4096$ images), ordered consistently with the channel order.

We provide an evaluative comparison of our method with other state-of-the-art methods on indoor, outdoor and wide-baseline datasets.
As our evaluations show, it is a viable alternative to methods involving explicit descriptors particularly with narrow baselines, achieving a similar pose estimation performance.

\section{Related Work}

Modern feature matching can be best understood by considering the sub-problems and their historical context.
Calculating dense correspondences for images used to be prohibitive, so instead early work concentrated on finding sparse sets of points that could be used for correspondence generation.
In a first interest point detector, \cite{Moravec80thesis} identified points which are explicitly {\it distinct} from the points that surround them---thus increasing the likelihood to be unambiguously matched in another image.
Subsequently, faster approximations for distinctiveness have been found, whether using first-order approximations \cite{Harris88, Shi94cvpr}, or convolutional filters~\cite{Lowe04ijcv,Bay08cviu}.
Alternatively, a detector can explicitly target a subset of distinctive points, such as corners and dots \cite{Rosten06eccv}.
All of these methods calculate a response for every pixel in the image, and the $n$ pixels with the largest response are selected as interest points.
This process typically involves non-maximum suppression in order to prevent directly neighboring points from being selected.

Once a set of interest points is extracted in the images, they need to be matched between each other to establish correspondences.
One can match points based directly on the surrounding image patches, but this is fragile to slight changes in illumination and viewpoint.
Instead, descriptors can be used, which are functions of patches and whose output is typically lower-dimensional, but invariant to some amount of illumination and viewpoint change, yet still distinctive enough to differ between the different points extracted in one image.
A popular class of traditional descriptors is histograms of gradients (HoG) \cite{Lowe04ijcv}.
Another example are binary descriptors, which are particularly efficient to calculate \cite{Calonder12pami, Leutenegger11iccv, Rublee11iccv}.

Most descriptors, however, are still sensitive to large affine transformations such as changes in scale and orientation.
Consequently, modern feature matchers use multi-scale detection \cite{Mikolajczyk04ijcv, Leutenegger11iccv}, and orientation estimation \cite{Lowe04ijcv, Rublee11iccv}.
A wide variety of traditional feature matching systems comprising these components exist, see the survey in \cite{Mukherjee15mva}.

Recent success of convolutional neural networks (CNNs), however, has led the community to revisit these systems and replacing their components with CNN-based methods.
For detection, the traditionally handcrafted \qmarks{featureness} responses can directly be replaced with a fully convolutional neural network.
Rather than just imitating traditional interest point detectors, CNN-based detectors can be trained to be invariant across different viewpoints \cite{Lenc16eccvw}, to present consistent ranking in the images in which they are extracted \cite{Savinov17cvpr} and to provide particularly sharp and thus unambiguous responses \cite{Zhang18cvpr}.
A majority of these methods is compared in \cite{Lenc18bmvc}.
CNNs are furthermore proven function approximators for image patches and thus well suited for descriptor calculations.
The output channels of a CNN can simply be interpreted as the coefficients of a descriptor \cite{Han15cvpr, Zagoruyko15cvpr, Simoserra15iccv, Mishchuk17nips, Loquercio17icra, Wei18cvpr, Keller18cvpr}.
A comparison of traditional and learned descriptors is provided in \cite{Schonberger17cvpr}.
While the results of comparing CNN-based methods to traditional methods do not yet suggest absolute superiority of CNN-based methods~\cite{Schonberger17cvpr}, an advantage of CNN-based methods is that they are malleable:
firstly, they can adapt to and learn from new data:
Consider an application where the type of environment is known beforehand---CNN-based methods can be trained to work particularly well on that particular type of environment.
Secondly, they can adapt to or be trained together with other components of a larger system.

Two recent systems that fully integrate CNN-based methods and do this kind of joint training are LF-Net \cite{Ono18nips} and SuperPoint \cite{Detone18cvprw}.
LF-Net~\cite{Ono18nips} builds on top of a previous method by the same authors, LIFT \cite{Yi16eccv}, the first such system, in which the method was trained on a set of pre-extracted patches.
\cite{Ono18nips}, instead, is trained in a self-supervised and more unconstrained manner, only requiring an image sequence with ground truth depths and poses.
Like \cite{Yi16eccv}, \cite{Ono18nips} uses separate CNNs for multi-scale interest point detection, feature orientation estimation and feature description.
In contrast, SuperPoint \cite{Detone18cvprw} only contains an interest point detector and a feature descriptor network, both of which are sharing several encoder layers.
It also does not explicitly express multi-scale detection, but rather trains multi-scale detection implicitly.
It is first pre-trained on labeled synthetic images, then fine tuned on artificially warped real images.
Both \cite{Ono18nips} and \cite{Detone18cvprw} still consider the traditional components of feature detection as separate functional units, even if the whole system is trained end-to-end.

In contrast, we offer a novel approach in which all components are subsumed into a single network.
Beside having the benefit that all components can be jointly trained (from scratch) and thus tailored to one another, we also get rid of explicit descriptors.
Instead, interest points are implicitly matched by the CNN output channel from which they originate.
In practice, this results in memory, computation and potentially data transmission savings, as descriptors do not need to be stored, matched or communicated any more.

There have been some previous attempts to significantly reduce the amount of data associated with descriptors.
In \cite{Tardioli15iros}, the authors replace descriptors with word identifiers of the corresponding visual word in a Bag-of-Words visual vocabulary \cite{Sivic03iccv}.
This can be used jointly with Bag-of-Words place recognition in order to facilitate multi-agent relative pose estimation with minimal data exchange.
In \cite{Lynen15rss}, the authors propose highly compressed maps for visual-inertial localization in which binary descriptors are projected down to as little as one byte.
In contrast, our approach circumvents the use of any explicit descriptor, by implicitly embedding a form of descriptor in the learned detection algorithm itself.

\section{System Overview}\label{sec:meth}

\begin{figure}
  \centering
  \includegraphics[width=.8\textwidth]{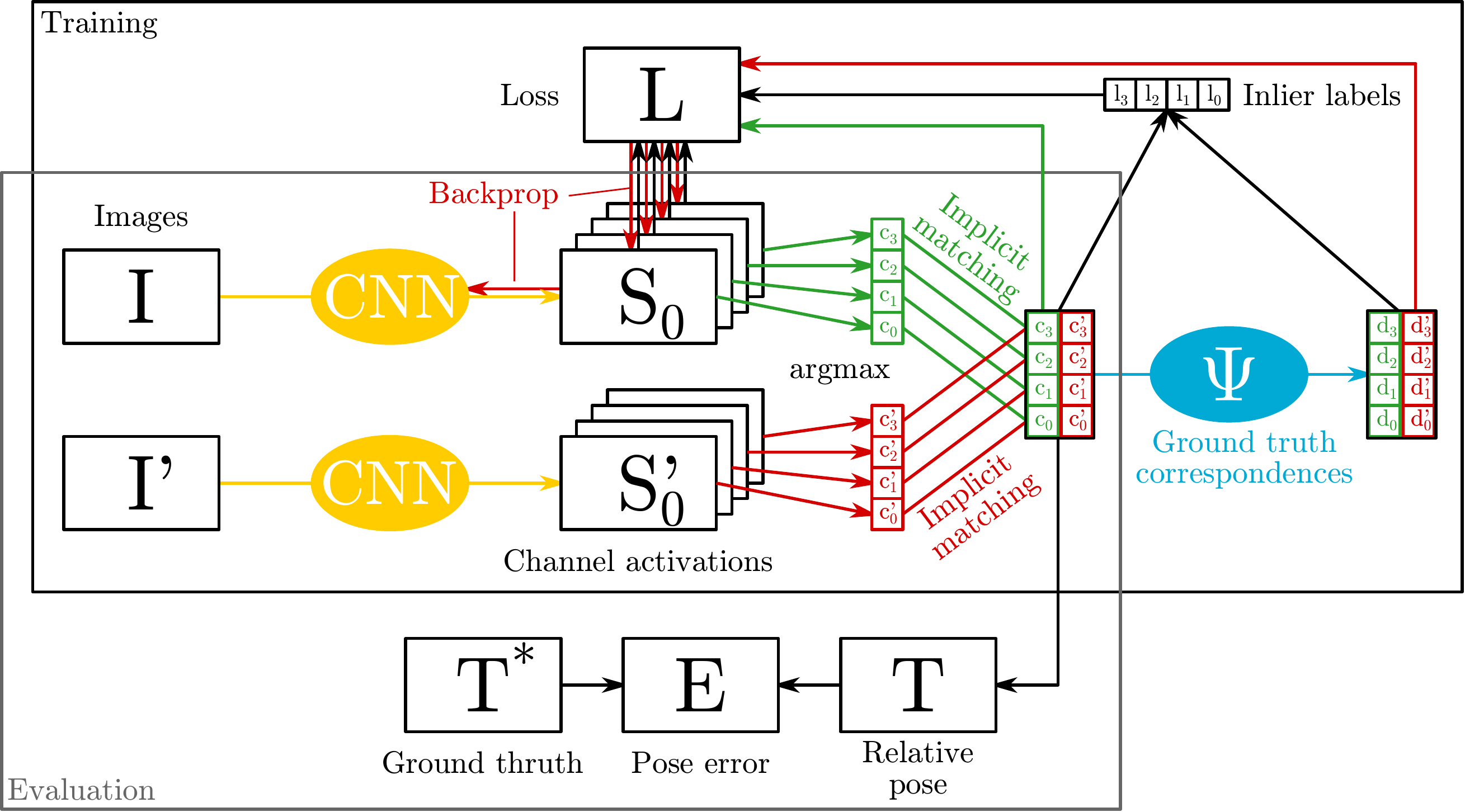}
  \vspace{-3mm}
  \caption{Overview of our approach.
  Given an image, a CNN computes $n$ activations, the argmax of which are considered the $n$ interest points.
  The interest points from two different images are matched by channel.
  During training, these matches are labeled as \emph{inliers}, \emph{outliers}, or \emph{unassigned} using ground truth correspondences.
  Our loss promotes inliers, penalizes outliers, and suppresses redundancy.
  For evaluation, the correspondences are used to compute a relative pose between two images which is compared against ground truth.}
  \label{fig:sys}
  \vspace{-5mm}
\end{figure}

In analogy to other state-of-the-art approaches for interest point detection, we employ a neural network to predict a per-pixel response from an input image.
But instead of only predicting a single output score for every pixel, we predict $n$ different activations, see \reffig{fig:sys}.
The neural network consists of $14$ layers of $3 \times 3$ convolutions with stride $1$ and leaky ReLU activations, except for the final activation, which is a sigmoid.
The first half layers output $64$ channels, the second half $128$.
From each final output channel $i$, we extract the argmax as $i$-th interest point with coordinates $\crd{i}$.
The key concept is that we then {\it implicitly} match the interest point from the same output channel across multiple frames.
This has the advantage of inherently solving the data association problem, without the need to use descriptors explicitly.
Formally, point $\crd{i}$ from image $\img$ is matched with point $\crd{i}'$ from image $\img'$.
At test time, a relative pose between both images can then be computed based on the corresponding interest point coordinates.

During training, an inlier determination module (\refsec{sec:inloutl}) processes the matches $\match$ and determines which of them are inliers.
This relies on ground truth correspondences $\corresp(\crd{i})$.
Interest points, correspondences and inlier labels shape mini-batches that contribute to the loss for a given training step as described in \refsec{sec:train}.
During evaluation and deployment, the inlier determination module is replaced with an application-specific geometric verifier, such as a perspective-n-point (PnP) localizer.
In our experiments, we evaluate our system with P3P \cite{Gao03pami} localization using RANSAC~\cite{Fischler81cacm}, which produces a relative pose estimate.
We compare this pose estimate to the ground truth relative pose to assess the viability of our method for visual pose estimation.

\subsection{Inlier and True Correspondence Determination}\label{sec:inloutl}

Our training methodology requires inlier labels and correspondences in order to calculate the loss.
Given a method $\corresp$ to calculate true correspondences between the two images provided in a training step, we label an interest point $\crd{i}$ as \emph{inlier} if the correspondence of the matched interest point from the other image $\corresp^{-1}(\crd{i}')$ is within $3px$ of the matched interest point in the other image and vice versa.
Otherwise, the match is either labeled an \emph{outlier} if the correspondence lies somewhere else within the respective image, or \emph{unassigned} if it is outside the image frame.
In some datasets, a method to compute ground truth correspondence is provided.
If such a method is not provided, correspondence can be calculated from ground truth depth and pose \cite{Savinov17cvpr}.
In case these are not provided either, the correspondence can be estimated using an SfM algorithm \cite{Ono18nips} given image sequences, or by using direct tracking such as KLT~\cite{Lucas81ijcai}.
With this only image regions with sufficient texture can be tracked, which is acceptable since image regions without texture are unlikely to be of interest.

\section{Training Methodology}\label{sec:train}

We use standard iterative training using Adam updates \cite{Kingma15iclr}.
In each iteration, a training sample, which consists of two images of the same scene is forwarded through the system to obtain a set of matches $\left\{\match, i \in \{0 , \ldots, n-1\}\right\}$ and an associated set of true correspondences $\{(\corrpt{i}, \corrpt{i}') = (\corresp(\crd{i}), \corresp^{-1}(\crd{i}'))\}$ according to \reffig{fig:sys}.
During training, the loss is only applied at these sparse locations.
In order to allow for efficient gradient backpropagation, patches are gathered from these locations.
In image $\img$, two mini-batches are formed:
one from stacking {\it interest point patches} $\patch{\crd{i}}$ centered around $\crd{i}$ and shaped according to the receptive field, $r \times r$, of a single pixel at the output.
The other from stacking {\it correspondence patches} $\patch{\corrpt{i}'}$ centered around the correspondences $\corrpt{i}'$ .
Both batches have a shape $ [n, r, r, 1] $.
The network transforms both of them into output tensors of shape $ [n, 1, 1, n] $.
The training loss is now applied to these tensors.
Since they are flat along the height and width dimensions, we can conceptualize them as square matrices along the batch and channel dimensions, and visualize how the loss is applied to them in \reftab{tab:loss} for a toy example with $n=3$.
\begin{table}
\centering
\begin{tabular}{|c||c||c|c|c||c|c|c|}
\hline
chn. & label & $\patch{\crd{0}}$ & $\patch{\crd{1}}$ & $\patch{\crd{2}}$ & $\patch{\corrpt{0}'}$ & $\patch{\corrpt{1}'}$ & $\patch{\corrpt{2}')}$ \\
\hline
0 & \cellcolor{green!50} inlier & \tlup &  &  &  &  & n/a \\
\hline
1 & \cellcolor{red!50} outlier & \tldown & \tldown &  &  & \tlup & n/a \\
\hline
2 & unass. & \tldown &  &  &  &  & n/a \\
\hline
\end{tabular}
\caption{Mini-batch toy example to illustrate losses.
For inliers, the activation of the maxima is strengthened while suppressing the activation in the other channels.
For outliers, the activation of the maxima is weakened while promoting the response of the true correspondence.}
\label{tab:loss}
\vspace{-2mm}
\end{table}
Note that the diagonal in the first tensor contains the responses of patch $\patch{\crd{i}}$ at channel $i$, which is the maximum value in channel $i$ and the value that caused $\crd{i}$ to be selected as interest point.
Similarly, the diagonal in the second tensor contains the responses that \emph{should} be the maximum in the given channel considering the correspondence from the interest point selected in the other image.
The training loss that is applied to these tensors has three components with their specific purpose:
\begin{itemize}
  \setlength\itemsep{-.4em}
 \item {\it Inlier reinforcement} reinforces interest points that are inliers in a given training sample, and suppresses interest points that are outliers.
 \item {\it Redundancy suppression} ensures that different channels do not converge to the same points.
 \item {\it Correspondence reinforcement} reinforces true correspondences of all points which are outliers in a given training sample.
\end{itemize}
The entire loss formulation is symmetrically applied to the other image $\img'$.

Let $\prob_{ij}$ be the scalar response of channel $j$ to patch $\patch{\crd{i}}$ and $\lab_i$ the inlier label of that patch.
Then, the inlier reinforcement loss is simply the cross-entropy loss according to that label, applied where $i = j$:
\vspace{-2mm}
\begin{align}
  \loss_{inl}(\prob_{ij}, \lab_i) = \left\{\begin{matrix} -\log(\prob_{ij}), \quad & \lab_i = \text{inlier}, i=j, \\ -\log(1 - \prob_{ij}), \quad & \lab_i = \text{outlier}, i=j, \\ 0\quad & \text{otherwise}. \end{matrix}\right.
\end{align}
This is the loss responsible for learning a high response for points that are likely to result in inliers.
In \reftab{tab:loss} the inlier reinforcement effect is observed for channel 0 on $\patch{\crd{0}}$ and the outlier suppression effect is present for channel 1 on $\patch{\crd{1}}$.

To prevent channels from converging to the same interest points, a loss is applied on inlier patches that suppresses the response on all channels, except the one which gave rise to the inlier:
\vspace{-2mm}
\begin{align}
  \loss_{red}(\prob_{ij}) = \left\{\begin{matrix} -\log(1 - \prob_{ij}), \quad & \lab_i = \text{inlier}, i\neq j, \\
  0 \quad & \text{otherwise}. \end{matrix}\right.
\end{align}
In \reftab{tab:loss} the redundancy suppression is present on channels 1 \& 2 on $\patch{\crd{0}}$.
We found that without redundancy suppression, all channels tend to converge to a single feature, all selecting the same interest point in every image.

Finally, we have found that our network does not converge with the above losses alone.
Thus, for channels with outliers, we promote $\prob_{ii}'$, the response of channel $i$ to patch $\patch{\corrpt{i}'}$:
\begin{align}
  \loss_{cor}(\prob_{ii}') = \left\{\begin{matrix} -\log(\prob_{ii}'), \quad & \lab_i = \text{outlier} \\ 0 \quad & \text{otherwise}. \end{matrix}\right.
\end{align}
In \reftab{tab:loss} the correspondence reinforcement can be observed in channel 1 on $\patch{\corrpt{1}'}$, which is the patch extracted at the correspondence of the maximum activation of the paired image.

\subsection{Training Pair Selection}\label{sec:pairsel}

In datasets such as HPatches \cite{Balntas17cvpr}, pairs are pre-selected.
For image sequences, we extract pairs as follows:
Given one image, points are densely sampled and subjected to KLT tracking for as far into subsequent images as possible.
A pair is then formed between this initial image and a random subsequent image in which at least a fraction $o$ of the initial points is still tracked.
$o$ thus reflects the minimum scene overlap between the two images.
The benefit of this method is that it can be applied to uncalibrated image sequences, while providing good guarantees regarding minimum scene overlap.
For training, we use $o=0.3$, for selecting pairs during evaluation $o=0.5$.

\section{Experiments}

We subject our method to evaluation and comparison to state-of-the-art on three datasets:
The viewpoint-variant part of the wide-baseline HPatches benchmark \cite{Balntas17cvpr, Lenc18bmvc}, sequence {\tt 00} of the outdoor autonomous driving dataset KITTI \cite{Geiger13ijrr} and sequence {\tt V1\_01} of the indoor drone dataset EuRoC \cite{Burri15ijrr}.
For the two sequences, $100$ image pairs are randomly selected using the method in \refsec{sec:pairsel}.
KITTI and EuRoC are stereo datasets, and we only use the left camera for the image pairs.
Rotation and translation differences within these pairs are plotted in the supplementary material.
For the HPatches benchmark, we train our network using the provided training split, while for KITTI and EuRoC, we train our network on a separate dataset, TUM mono \cite{Engel17pami}, where we use the rectified images of sequences {\tt 01}, {\tt 02}, {\tt 03} (indoors) and {\tt 48}, {\tt 49}, {\tt 50} (outdoors).
We found that adding sequences {\tt 04} to {\tt 47} does not result in better performance.
For the sequences, we use KITTI {\tt 05} as validation dataset.
Ground truth correspondences are given by ground truth homographies in the HPatches training data.
In TUM mono, they are instead established using KLT as discussed in \refsec{sec:inloutl}.

We compare our method to SIFT~\cite{Lowe04ijcv}, SURF~\cite{Bay08cviu}, ORB~\cite{Rublee11iccv}, LF-Net~\cite{Ono18nips} and SuperPoint~\cite{Detone18cvprw}.
For SIFT, SURF and ORB, we use the OpenCV implementation, while for LF-Net and Superpoint, we use the publicly available code and pre-trained weights.
All baselines are evaluated both at $128$ interest points, like our method, and at a more native interest point count of $500$.

\subsection{Matching score}

\begin{figure}
  \centering
  \begin{tabular}{ccc}
  \includegraphics[trim=0 0 35 0,clip,width=.3375\linewidth]{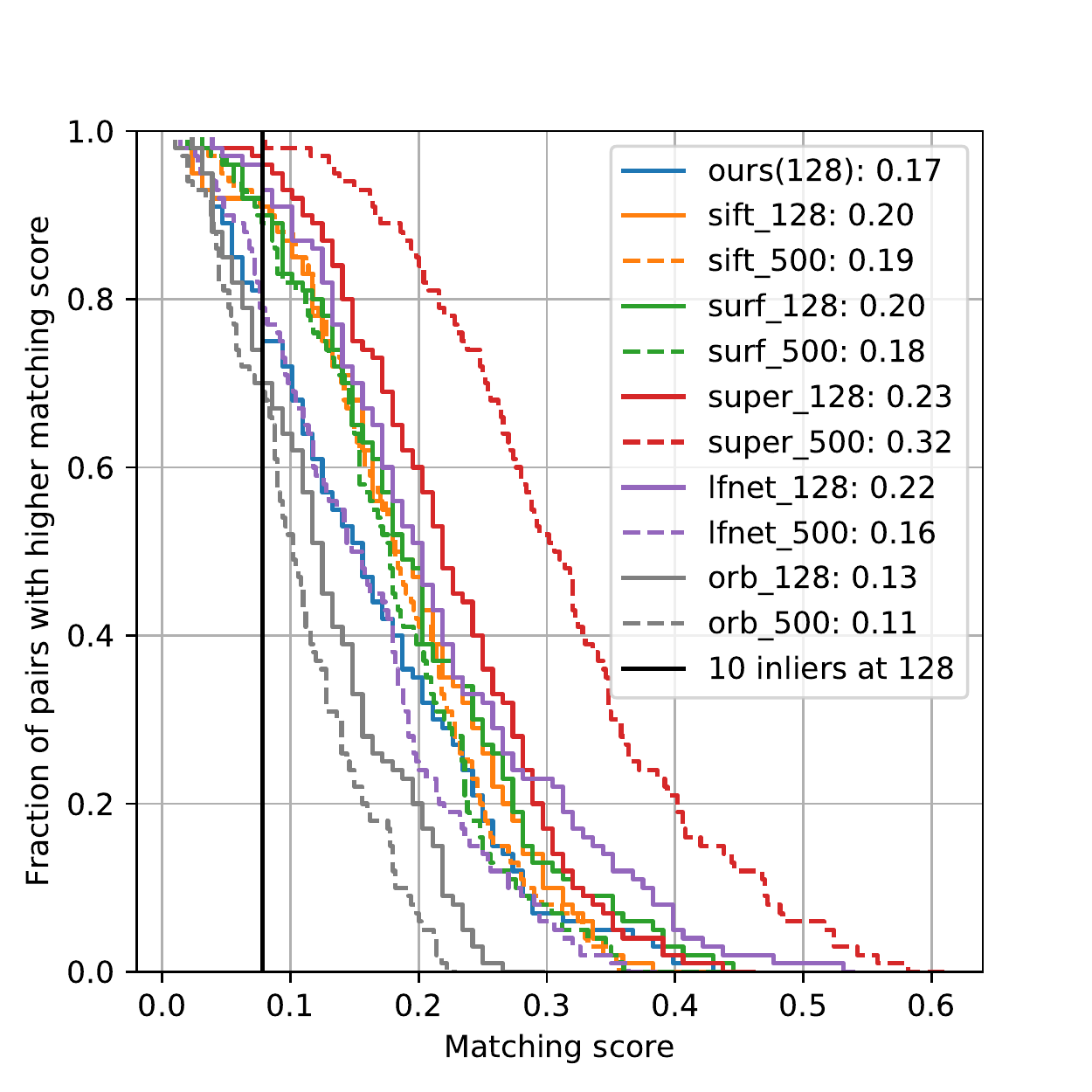}&
  \includegraphics[trim=40 0 35 0,clip,width=.296\linewidth]{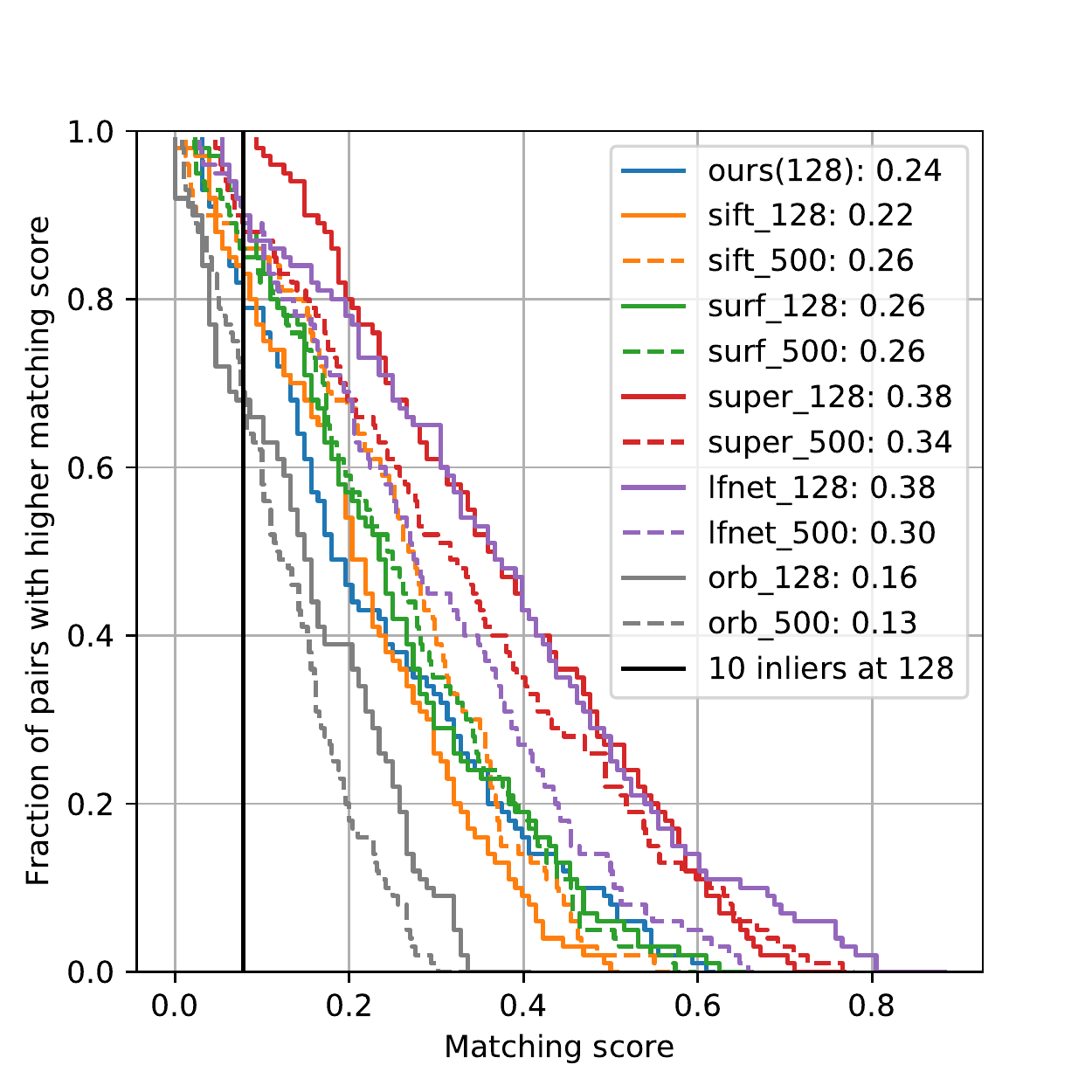}&
  \includegraphics[trim=40 0 35 0,clip,width=.296\linewidth]{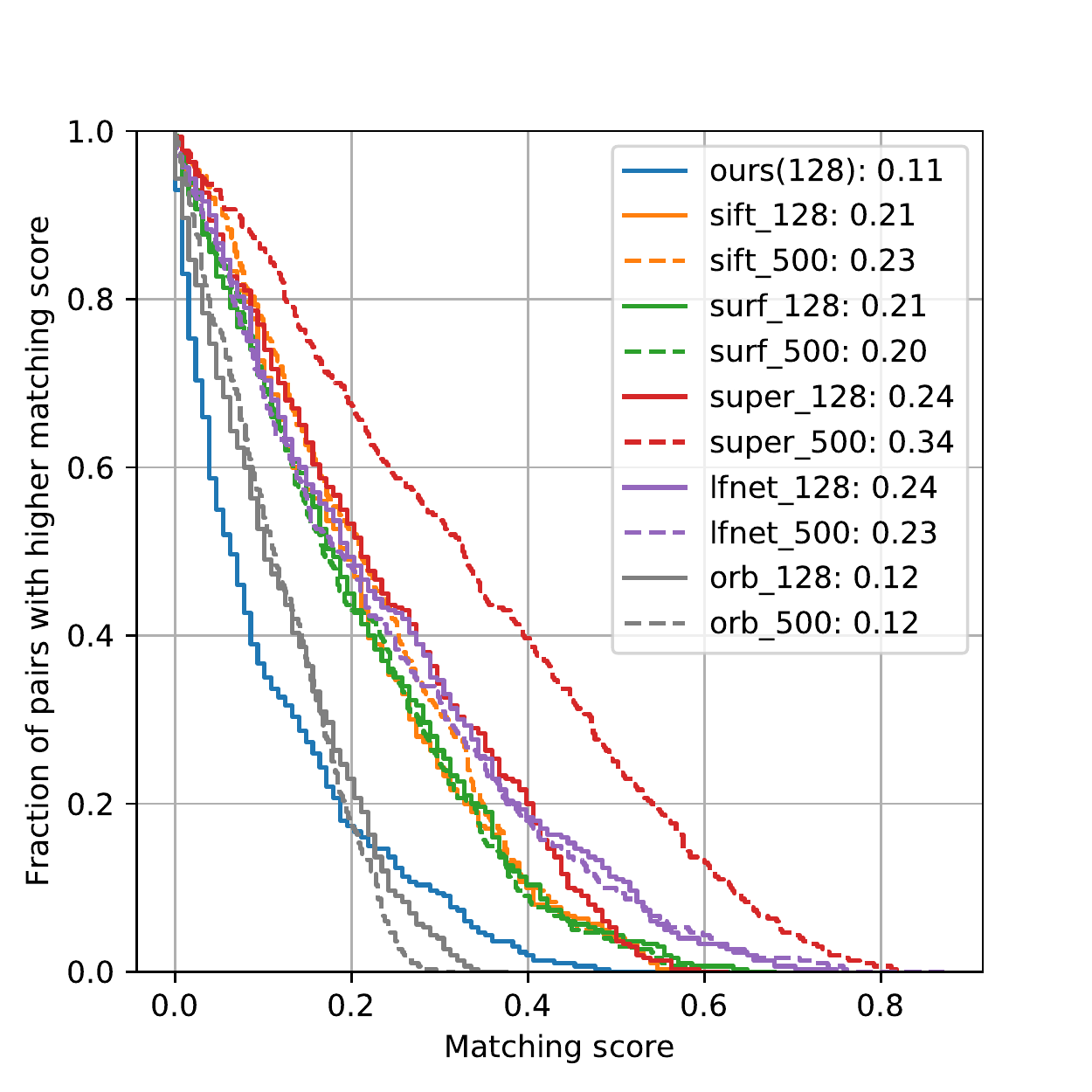}\\
  (a) KITTI & (b) Euroc & (c) HPatches
  \end{tabular}
  \caption{Matching score distributions (higher is better).}
  \label{fig:mscore}
  \vspace{-5mm}
\end{figure}

For all datasets, we evaluate matching score, which is the fraction of inlier matches among all matches.
In HPatches, the ground truth homography is used to distinguish inlier from outlier matches.
In KITTI and EuRoC we instead use the inliers that are determined during pose estimation, see \refsec{sec:posest}.
As a consequence, the matching score here more closely reflects the matching score that is achieved in practice when doing relative pose estimation, whether in SLAM or in map localization.
\reffig{fig:mscore} shows the matching score obtained in the three datasets.
Both on KITTI and on EuRoC, our method performs slightly worse than all baselines except ORB, especially if the latter are permitted to extract more points than our method.
Note, however, that SURF, SIFT, SuperPoint and LF-NET use $64, 128, 256$ and $256$ floating points to describe each interest point, respectively, in addition to point locations.
Instead, our method represents a visual location using only $128$ properly ordered point locations.
The results on HPatches show that our method is still not very well suited to the significant viewpoint and scale changes present in that dataset.
Nevertheless, we believe that implicit interest point matching opens a new research direction in terms of more closely integrated detector training.
Future work to address strong viewpoint and scale changes could include considerations such as scale and rotation invariance, whether by modeling inside the neural network \cite{Ono18nips} or by data augmentation and curricular learning \cite{Detone18cvprw}.

\subsection{Pose Estimation Accuracy}\label{sec:posest}

\begin{figure}
  \centering
  \begin{tabular}{ccc}
  \includegraphics[trim=0 0 5 0,clip,width=.352\linewidth]{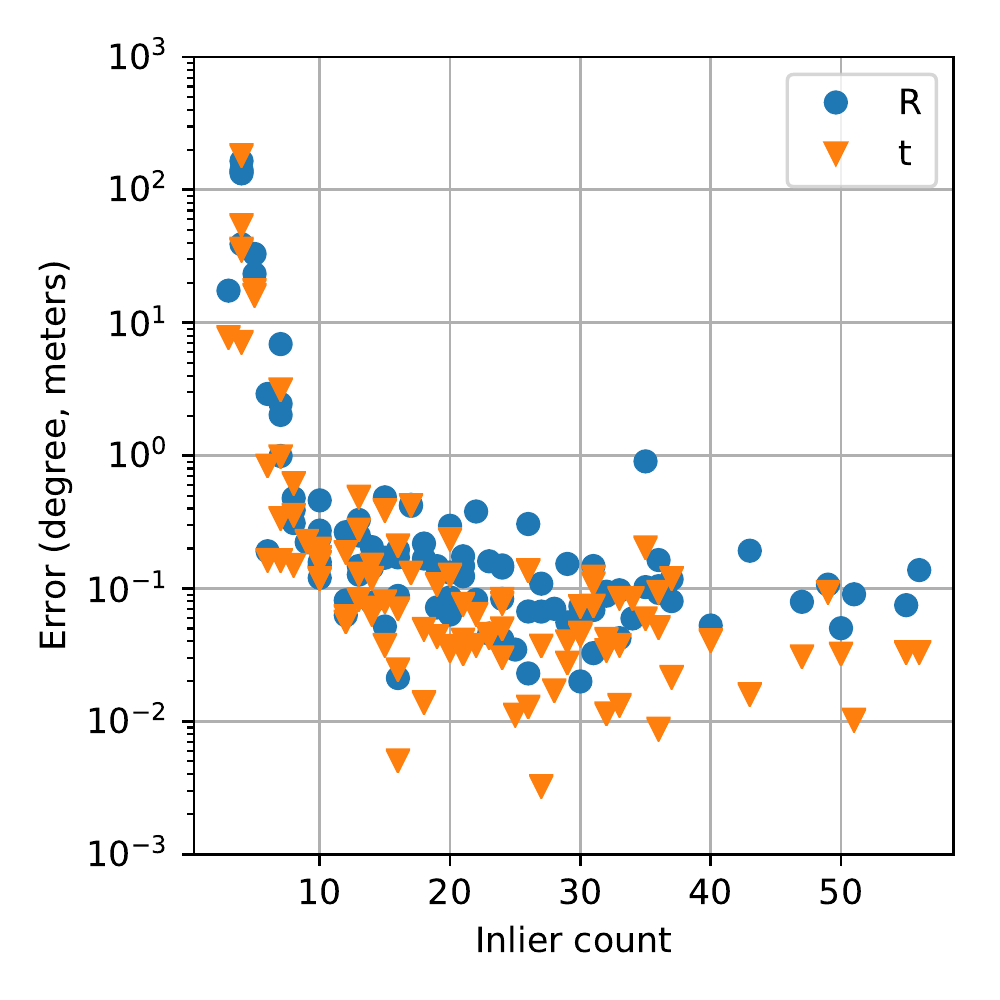}&
  \includegraphics[trim=50 0 5 0,clip,width=.29\linewidth]{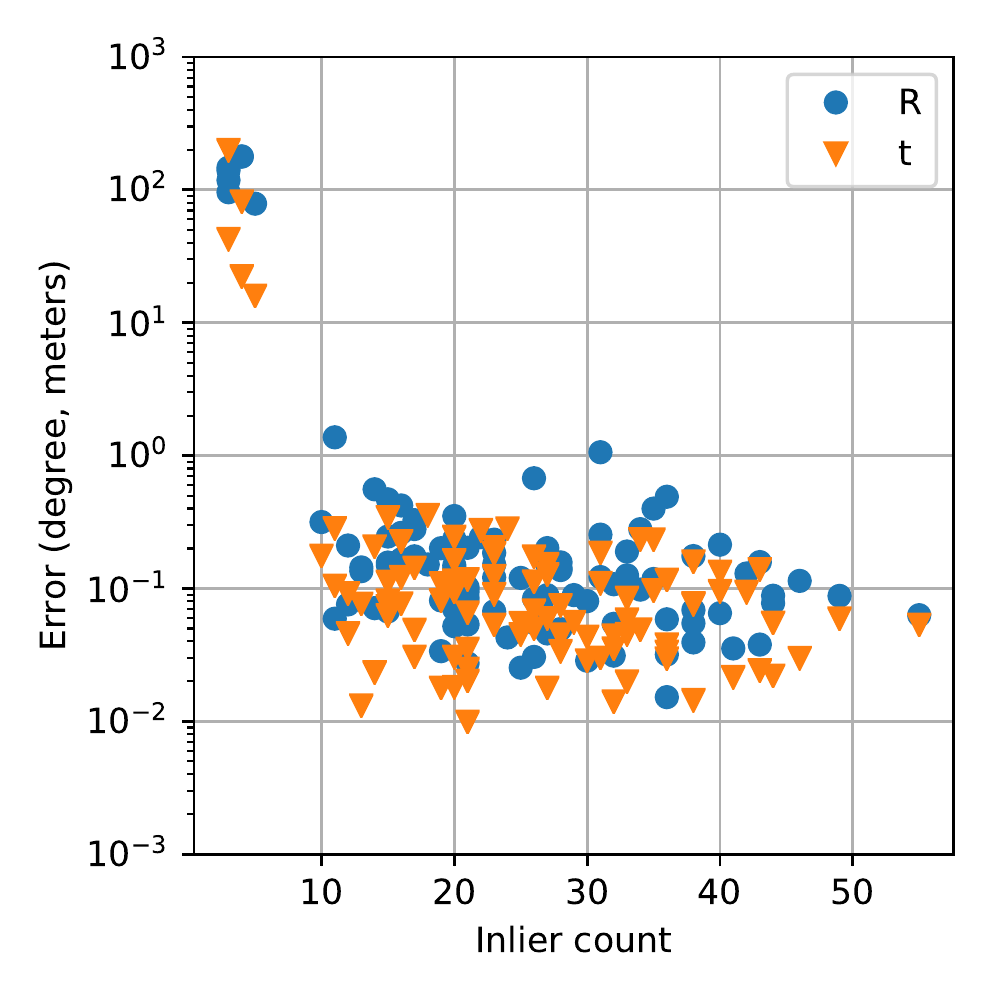}&
  \includegraphics[trim=50 0 5 0,clip,width=.29\linewidth]{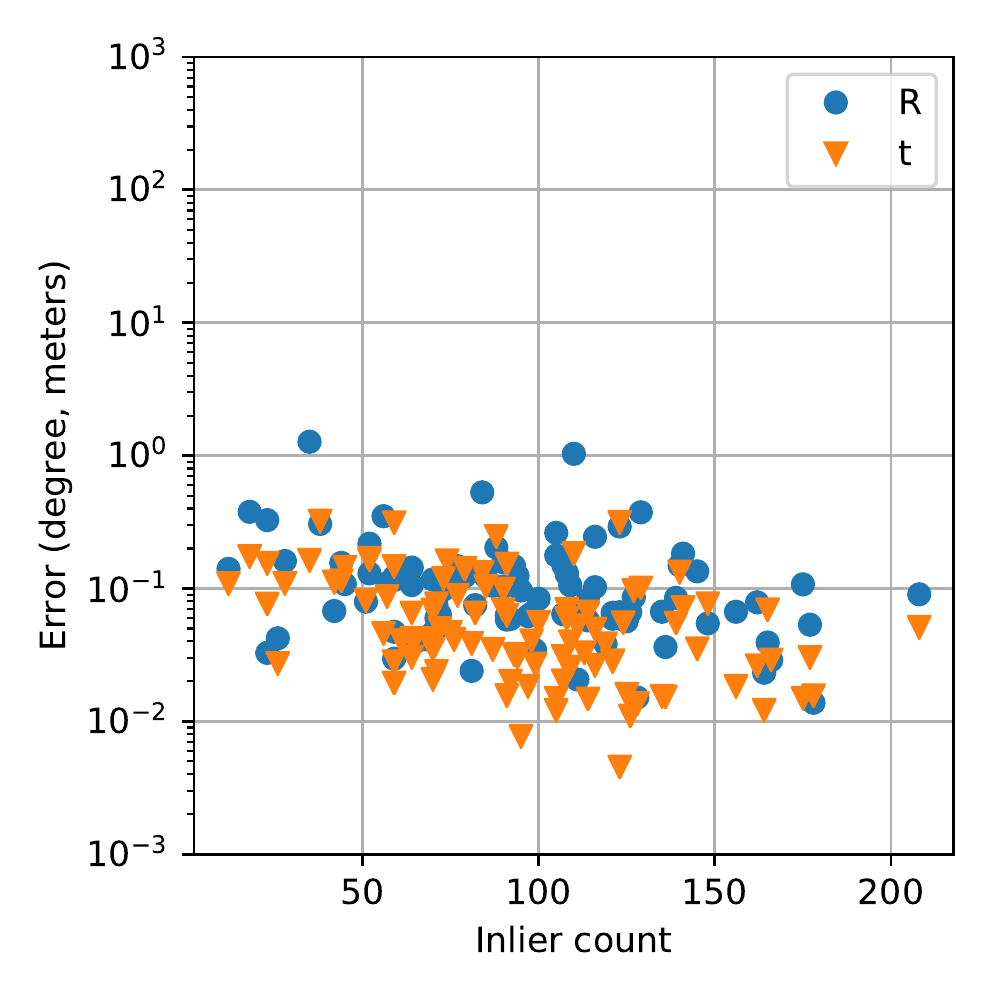}\\
  (a) ours & (b) SIFT, 128 points & (c) SIFT, 500 points
  \end{tabular}
  \caption{Pose error versus inlier count on KITTI.
  R: rotation error, t: translation error.}
  \label{fig:rt}
  \vspace{-5mm}
\end{figure}

To understand how matching score translates to pose estimation accuracy, we compare the pose estimated using our method and SIFT with the ground truth relative pose on both KITTI and EuRoC.
Both in SLAM and map localization absolute relative pose is typically obtained from interest points matched to 3D point locations by P3P localization~\cite{Gao03pami} with RANSAC~\cite{Fischler81cacm}.
To obtain 3D point locations in one of the two images, we use epipolar stereo matching of the interest points extracted using our method or a baseline with the corresponding image from the right camera.
Rotation and translation error are measured with the geodesic (angle in angle-axis) and euclidean distances.
In \reffig{fig:rt} these errors are compared to the inlier count for each pair in the KITTI testing set, using both our interest points and SIFT.
The same plot for EuRoC can be found in the supplementary material.
We can see that both with our method and with SIFT, an inlier count above 10 indicates a good relative pose (rotation error below $1^{\circ}$, translation error below $30$cm).
Estimates get slightly better up to an inlier count of $20$, beyond which they do not improve by much, even if much more interest points are extracted.
We transfer this insight to \reffig{fig:mscore}, where we indicate the matching score corresponding to $10$ inliers with a vertical line.
From this, we can see that our method results in good relative poses in $80\%$ of the tested image pairs.

\subsection{Other results}

\begin{figure}
  \centering
  \begin{tabular}{cccc}
  (a)\hspace{-5mm} & \multirow{2}{*}{\includegraphics[width=.46\linewidth]{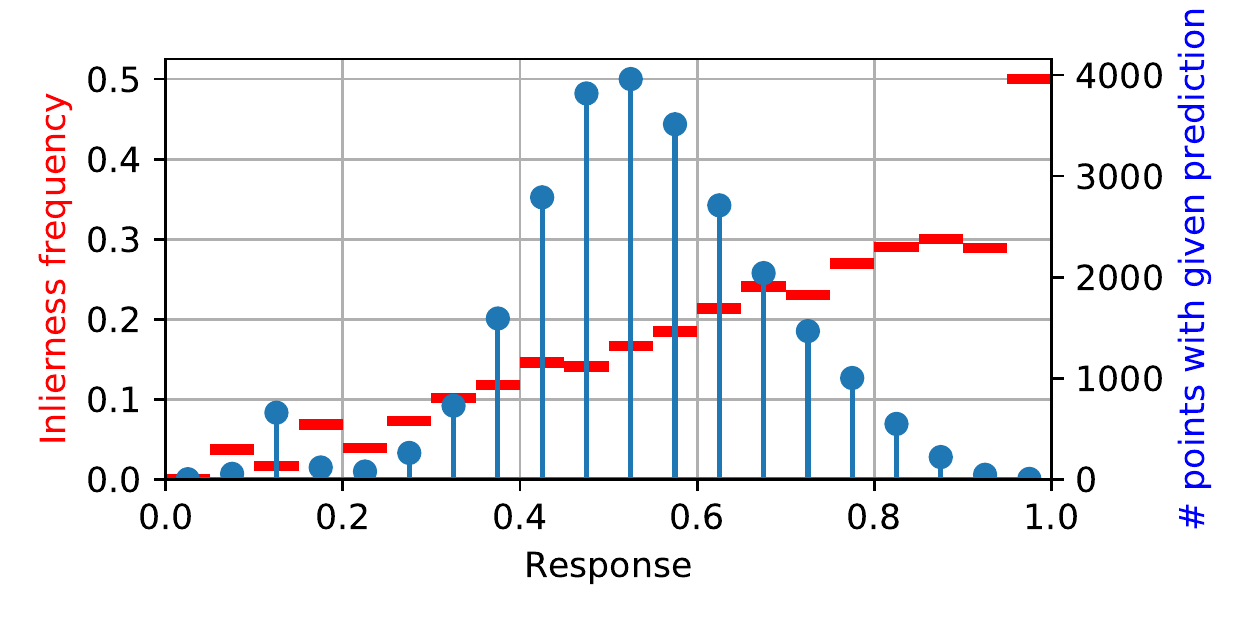}\hspace{-5mm}} &
  (b)\hspace{-5mm} & \multirow{2}{*}{\includegraphics[width=.46\linewidth]{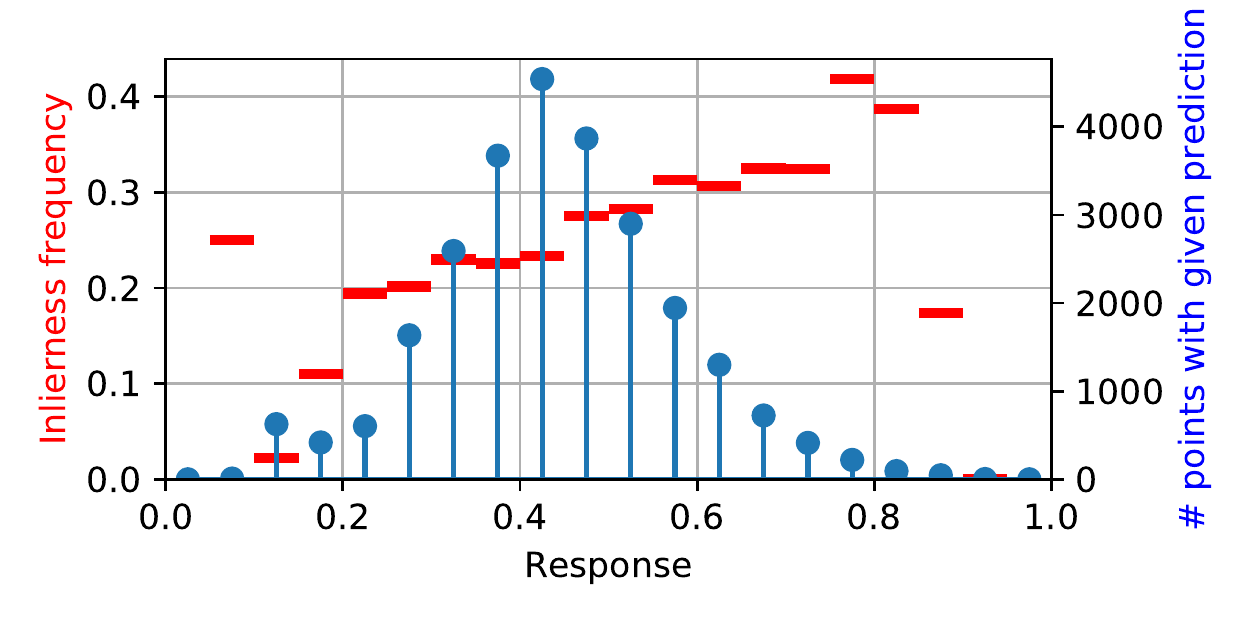}\hspace{-5mm}} \\
  \vspace{19mm} & & & \\
  (c)\hspace{-5mm} & \multirow{2}{*}{\includegraphics[width=.46\linewidth]{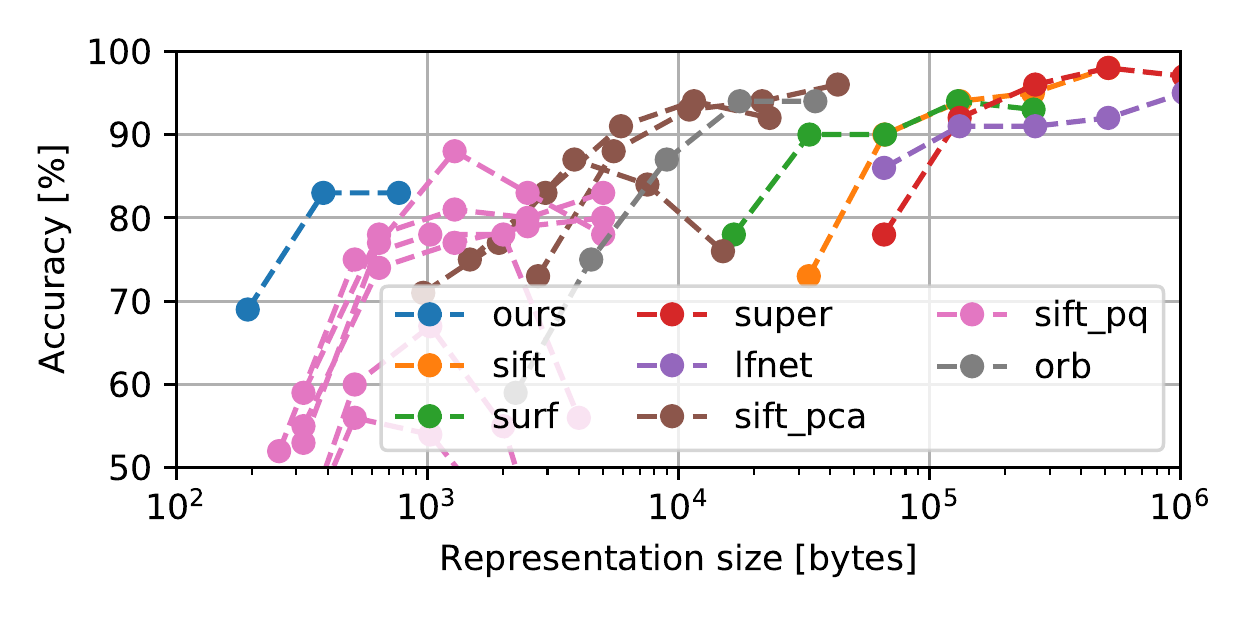}\hspace{-5mm}} &
  (d)\hspace{-5mm} & \multirow{2}{*}{\includegraphics[width=.46\linewidth]{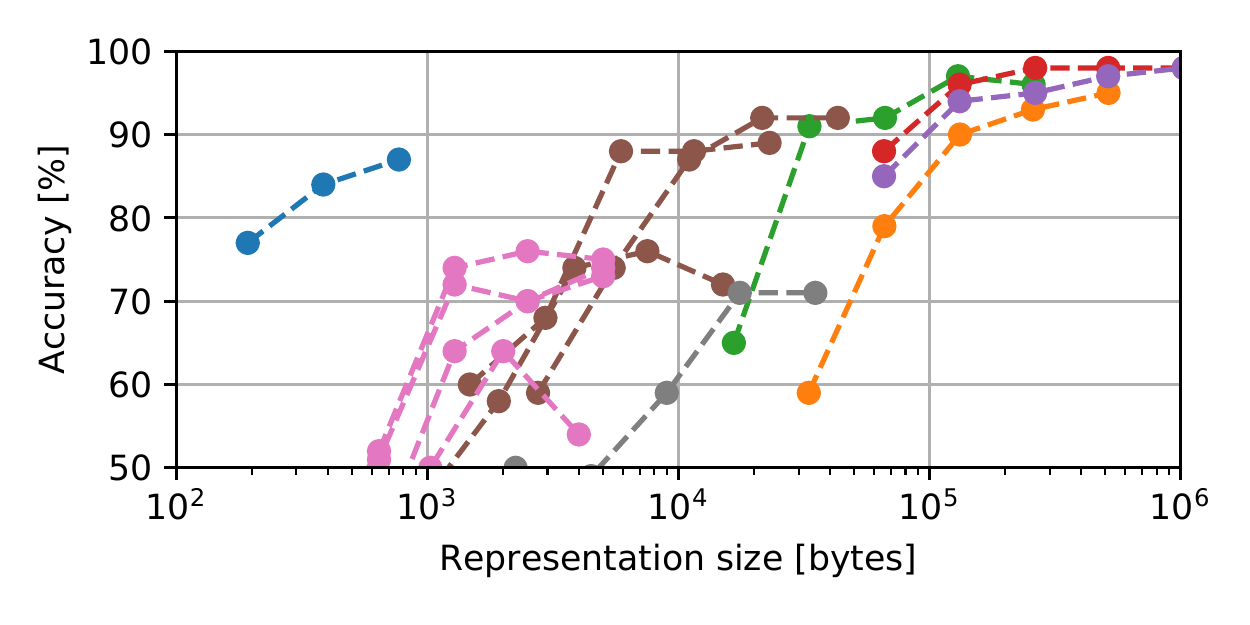}\hspace{-5mm}}\\
  \vspace{19mm} & & & \\
  \end{tabular}
  \caption{
  (a), (b) Empiric \qmarks{inlierness} frequency versus response, and histogram of interest point responses.
  Inlierness frequency is calculated per histogram bin by dividing the amount of inliers within a response range by the total amount of interest points within that range.
  (c), (d) Accuracy versus representation size for our method and baselines, including compression schemes for SIFT.
  Note the logarithmic scale.
  Left: KITTI, right: EuRoC.}
  \label{fig:moar}
  \vspace{-5mm}
\end{figure}

Some additional insights are shown in \reffig{fig:moar}.
The parts (a) and (b) show how the probability of a point being an inlier is correlated with the response of the corresponding channel at that point, using our network.
As can be seen, there is a good correlation, which indicates that the response has some predictive power regarding the probability of a point resulting in an inlier.
This could potentially be exploited, for example in RANSAC model sampling.
In (c) and (d), we show how pose estimation accuracy compares to the amount of data needed to represent a visual frame for our method, the previously used baselines and two compression methods applied to SIFT descriptors.
Here, accuracy represents the fraction of pose estimates with rotation and translation errors below $1^{\circ}$ and $30$cm (KITTI) and $3^{\circ}$ and $10$cm (EuRoC).
The two compression methods are principle component analysis (PCA) projection of SIFT descriptors and product quantization \cite{Jegou11pami, Lynen15rss}, see the supplementary material for details.
For our method, we evaluate networks with output channel counts $\in \{64, 128, 256\}$.
For all other methods, we evaluate $\{64, 128, 256, 500, 1000\}$ interest points per image.
As can be seen, our method outperforms baselines in the trade-off between accuracy and representation size.

\section{Conclusion}
In this paper, we have introduced a descriptor-free approach for detection and matching of sparse visual features.
We rely on a convolutional neural network to predict multiple activation layers and we define the location of the maximal response in each layer to be an interest point.
The key novelty is that instead of relying on descriptors for matching, interest points are uniquely associated to the activation layer they are extracted from.
This setup allows us to train the traditionally modularized interest point detection, description, and matching processes jointly in a simple setup, while at the same time getting rid of the requirement for explicit descriptors.
Without descriptors, visual features can be stored, communicated and matched at a highly reduced cost.
For this system, we have devised a self-supervised training methodology that reinforces interest points that result in inliers, ensures that each channel specializes on different features, and uses ground truth correspondences to ensure that channels not resulting in inliers find good candidate features.
This training can be achieved with uncalibrated image sequences.
Albeit achieving slightly lower matching scores when compared to other approaches that do use descriptors, we demonstrated the applicability of our descriptor-free approach in a visual pose estimation setup.

\section{Acknowledgments}

This work was supported by the National Centre of Competence in Research (NCCR) Robotics through the Swiss National Science Foundation and the SNSF-ERC Starting Grant.
The Titan Xp used for this research was donated by the NVIDIA Corporation.

\clearpage

\bibliography{all}

\begin{thebibliography}{41}
\providecommand{\natexlab}[1]{#1}
\providecommand{\url}[1]{\texttt{#1}}
\expandafter\ifx\csname urlstyle\endcsname\relax
  \providecommand{\doi}[1]{doi: #1}\else
  \providecommand{\doi}{doi: \begingroup \urlstyle{rm}\Url}\fi

\bibitem[Balntas et~al.(2017)Balntas, Lenc, Vedaldi, and
  Mikolajczyk]{Balntas17cvpr}
Vassileios Balntas, Karel Lenc, Andrea Vedaldi, and Krystian Mikolajczyk.
\newblock {HPatches}: A benchmark and evaluation of handcrafted and learned
  local descriptors.
\newblock In \emph{{IEEE} Conf. Comput. Vis. Pattern Recog. (CVPR)}, 2017.

\bibitem[Bay et~al.(2008)Bay, Ess, Tuytelaars, and Van~Gool]{Bay08cviu}
H.~Bay, A.~Ess, T.~Tuytelaars, and L.~Van~Gool.
\newblock {SURF}: Speeded up robust features.
\newblock \emph{Comput. Vis. Image. Und.}, 110\penalty0 (3):\penalty0 346--359,
  2008.
\newblock \doi{10.1016/j.cviu.2007.09.014}.

\bibitem[Burri et~al.(2015)Burri, Nikolic, Gohl, Schneider, Rehder, Omari,
  Achtelik, and Siegwart]{Burri15ijrr}
Michael Burri, Janosch Nikolic, Pascal Gohl, Thomas Schneider, Joern Rehder,
  Sammy Omari, Markus~W. Achtelik, and Roland Siegwart.
\newblock The {EuRoC} micro aerial vehicle datasets.
\newblock \emph{Int. J. Robot. Research}, 35\penalty0 (10):\penalty0
  1157--1163, 2015.
\newblock \doi{10.1177/0278364915620033}.

\bibitem[Calonder et~al.(2012)Calonder, Lepetit, Ozuysal, Trzcinski, Strecha,
  and Fua]{Calonder12pami}
M.~Calonder, V.~Lepetit, M.~Ozuysal, T.~Trzcinski, C.~Strecha, and P.~Fua.
\newblock {BRIEF}: Computing a local binary descriptor very fast.
\newblock \emph{{IEEE} Trans. Pattern Anal. Mach. Intell.}, 34\penalty0
  (7):\penalty0 1281--1298, 2012.

\bibitem[DeTone et~al.(2018)DeTone, Malisiewicz, and Rabinovich]{Detone18cvprw}
Daniel DeTone, Tomasz Malisiewicz, and Andrew Rabinovich.
\newblock {SuperPoint}: Self-supervised interest point detection and
  description.
\newblock In \emph{{IEEE} Conf. Comput. Vis. Pattern Recog. Workshops (CVPRW)},
  2018.

\bibitem[Engel et~al.(2018)Engel, Koltun, and Cremers]{Engel17pami}
Jakob Engel, Vladlen Koltun, and Daniel Cremers.
\newblock {D}irect {S}parse {O}dometry.
\newblock \emph{{IEEE} Trans. Pattern Anal. Mach. Intell.}, 40\penalty0
  (3):\penalty0 611--625, March 2018.
\newblock \doi{10.1109/TPAMI.2017.2658577}.

\bibitem[Farneb{\"a}ck(2003)]{Farnebaeck03scia}
Gunnar Farneb{\"a}ck.
\newblock Two-frame motion estimation based on polynomial expansion.
\newblock In \emph{Scandinavian Conf. on Im. Analysis ({SCIA})}, pages
  363--370, 2003.

\bibitem[Fischler and Bolles(1981)]{Fischler81cacm}
Martin~A. Fischler and Robert~C. Bolles.
\newblock Random sample consensus: a paradigm for model fitting with
  applications to image analysis and automated cartography.
\newblock \emph{Commun. {ACM}}, 24\penalty0 (6):\penalty0 381--395, 1981.
\newblock \doi{10.1145/358669.358692}.

\bibitem[Gao et~al.(2003)Gao, Hou, Tang, and Cheng]{Gao03pami}
Xiao-Shan Gao, Xiao-Rong Hou, Jianliang Tang, and Hang-Fei Cheng.
\newblock Complete solution classification for the perspective-three-point
  problem.
\newblock \emph{{IEEE} Trans. Pattern Anal. Mach. Intell.}, 25\penalty0
  (8):\penalty0 930--943, August 2003.
\newblock \doi{10.1109/TPAMI.2003.1217599}.

\bibitem[Geiger et~al.(2013)Geiger, Lenz, Stiller, and Urtasun]{Geiger13ijrr}
Andreas Geiger, Philip Lenz, Christoph Stiller, and Raquel Urtasun.
\newblock Vision meets robotics: The {KITTI} dataset.
\newblock \emph{Int. J. Robot. Research}, 32\penalty0 (11):\penalty0
  1231--1237, 2013.
\newblock \doi{10.1177/0278364913491297}.

\bibitem[Han et~al.(2015)Han, Leung, Jia, Sukthankar, and Berg]{Han15cvpr}
Xufeng Han, Thomas Leung, Yangqing Jia, Rahul Sukthankar, and Alexander~C.
  Berg.
\newblock {MatchNet}: Unifying feature and metric learning for patch-based
  matching.
\newblock In \emph{{IEEE} Conf. Comput. Vis. Pattern Recog. (CVPR)}, 2015.

\bibitem[Harris and Stephens(1988)]{Harris88}
Chris Harris and Mike Stephens.
\newblock A combined corner and edge detector.
\newblock In \emph{Proc. Fourth Alvey Vision Conf.}, volume~15, pages 147--151,
  1988.
\newblock \doi{10.5244/C.2.23}.

\bibitem[Horn and Schunck(1981)]{Horn81ai}
Berthold~K.P. Horn and Brian~G. Schunck.
\newblock Determining optical flow.
\newblock \emph{J. Artificial Intell.}, 17\penalty0 (1):\penalty0 185 -- 203,
  1981.
\newblock \doi{10.1016/0004-3702(81)90024-2}.

\bibitem[{Jegou} et~al.(2011){Jegou}, {Douze}, and {Schmid}]{Jegou11pami}
Herve {Jegou}, Matthijs {Douze}, and Cordelia {Schmid}.
\newblock Product quantization for nearest neighbor search.
\newblock \emph{{IEEE} Trans. Pattern Anal. Mach. Intell.}, 33\penalty0
  (1):\penalty0 117--128, January 2011.
\newblock \doi{10.1109/TPAMI.2010.57}.

\bibitem[Keller et~al.(2018)Keller, Chen, Maffra, Schmuck, and
  Chli]{Keller18cvpr}
Michel Keller, Zetao Chen, Fabiola Maffra, Patrik Schmuck, and Margarita Chli.
\newblock Learning deep descriptors with scale-aware triplet networks.
\newblock In \emph{{IEEE} Conf. Comput. Vis. Pattern Recog. (CVPR)}, 2018.

\bibitem[Kingma and Ba(2015)]{Kingma15iclr}
Diederik~P. Kingma and Jimmy~L. Ba.
\newblock Adam: A method for stochastic optimization.
\newblock In \emph{Int. Conf. Learn. Representations ({ICLR})}, 2015.

\bibitem[Lenc and Vedaldi(2016)]{Lenc16eccvw}
Karel Lenc and Andrea Vedaldi.
\newblock Learning covariant feature detectors.
\newblock In \emph{Eur. Conf. Comput. Vis. Workshops (ECCVW)}, pages 100--117,
  2016.

\bibitem[Lenc and Vedaldi(2018)]{Lenc18bmvc}
Karel Lenc and Andrea Vedaldi.
\newblock Large scale evaluation of local image feature detectors on homography
  datasets.
\newblock In \emph{British Mach. Vis. Conf. (BMVC)}, 2018.

\bibitem[Leutenegger et~al.(2011)Leutenegger, Chli, and
  Siegwart]{Leutenegger11iccv}
S.~Leutenegger, M.~Chli, and R.Y. Siegwart.
\newblock {BRISK}: Binary robust invariant scalable keypoints.
\newblock In \emph{Int. Conf. Comput. Vis. (ICCV)}, pages 2548--2555, 2011.
\newblock \doi{10.1109/ICCV.2011.6126542}.

\bibitem[Loquercio et~al.(2017)Loquercio, Dymczyk, Zeisl, Lynen, Gilitschenski,
  and Siegwart]{Loquercio17icra}
Antonio Loquercio, Marcin Dymczyk, Bernhard Zeisl, Simon Lynen, Igot
  Gilitschenski, and Roland Siegwart.
\newblock Efficient descriptor learning for large scale localization.
\newblock In \emph{{IEEE} Int. Conf. Robot. Autom. (ICRA)}, pages 3170--3177,
  2017.
\newblock \doi{10.1109/ICRA.2017.7989359}.

\bibitem[Lowe(2004)]{Lowe04ijcv}
David~G. Lowe.
\newblock Distinctive image features from scale-invariant keypoints.
\newblock \emph{Int. J. Comput. Vis.}, 60\penalty0 (2):\penalty0 91--110,
  November 2004.
\newblock \doi{10.1023/B:VISI.0000029664.99615.94}.

\bibitem[Lucas and Kanade(1981)]{Lucas81ijcai}
Bruce~D. Lucas and Takeo Kanade.
\newblock An iterative image registration technique with an application to
  stereo vision.
\newblock In \emph{Int. Joint Conf. Artificial Intell. (IJCAI)}, pages
  674--679, 1981.

\bibitem[Lynen et~al.(2015)Lynen, Sattler, Bosse, Hesch, Pollefeys, and
  Siegwart]{Lynen15rss}
Simon Lynen, Torsten Sattler, Michael Bosse, Joel Hesch, Marc Pollefeys, and
  Roland Siegwart.
\newblock Get out of my lab: Large-scale, real-time visual-inertial
  localization.
\newblock In \emph{Robotics: Science and Systems (RSS)}, 2015.
\newblock \doi{10.15607/RSS.2015.XI.037}.

\bibitem[Mikolajczyk and Schmid(2004)]{Mikolajczyk04ijcv}
K.~Mikolajczyk and C.~Schmid.
\newblock Scale \& affine invariant interest point detectors.
\newblock \emph{Int. J. Comput. Vis.}, 60\penalty0 (1):\penalty0 63--86, 2004.
\newblock \doi{10.1023/B:VISI.0000027790.02288.f2}.

\bibitem[Mishchuk et~al.(2017)Mishchuk, Mishkin, Radenovic, and
  Matas]{Mishchuk17nips}
Anastasiia Mishchuk, Dmytro Mishkin, Filip Radenovic, and Ji\u{r}i Matas.
\newblock Working hard to know your neighbors margins: Local descriptor
  learning loss.
\newblock In \emph{Conf. Neural Inf. Process. Syst. (NeurIPS)}, pages
  4826--4837. 2017.

\bibitem[Moravec(1980)]{Moravec80thesis}
H.~P. Moravec.
\newblock \emph{Obstacle Avoidance and Navigation in the Real World by Seeing
  Robot Rover}.
\newblock PhD thesis, Carnegie-Mellon University, Pittsburgh, Pennsylvania,
  September 1980.

\bibitem[Mukherjee et~al.(2015)Mukherjee, Wu, and Wang]{Mukherjee15mva}
Dibyendu Mukherjee, Q.~M.~Jonathan Wu, and Guanghui Wang.
\newblock A comparative experimental study of image feature detectors and
  descriptors.
\newblock \emph{Mach. Vis. and Applications}, 26\penalty0 (4):\penalty0
  443--466, May 2015.
\newblock \doi{10.1007/s00138-015-0679-9}.

\bibitem[Ono et~al.(2018)Ono, Trulls, Fua, and Yi]{Ono18nips}
Yuki Ono, Eduard Trulls, Pascal Fua, and Kwang~Moo Yi.
\newblock {LF-Net}: Learning local features from images.
\newblock In \emph{Conf. Neural Inf. Process. Syst. (NeurIPS)}, pages
  6234--6244, 2018.

\bibitem[Rocco et~al.(2018)Rocco, Cimpoi, Arandjelovi\'{c}, Torii, Pajdla, and
  Sivic]{Rocco18nips}
Ignacio Rocco, Mircea Cimpoi, Relja Arandjelovi\'{c}, Akihiko Torii, Tomas
  Pajdla, and Josef Sivic.
\newblock Neighbourhood consensus networks.
\newblock In \emph{Conf. Neural Inf. Process. Syst. (NeurIPS)}, pages
  1651--1662, 2018.

\bibitem[Rosten and Drummond(2006)]{Rosten06eccv}
Edward Rosten and Tom Drummond.
\newblock Machine learning for high-speed corner detection.
\newblock In \emph{Eur. Conf. Comput. Vis. (ECCV)}, pages 430--443, 2006.
\newblock \doi{10.1007/11744023_34}.

\bibitem[Rublee et~al.(2011)Rublee, Rabaud, Konolige, and
  Bradski]{Rublee11iccv}
E.~Rublee, V.~Rabaud, K.~Konolige, and G.~Bradski.
\newblock {ORB}: An efficient alternative to {SIFT} or {SURF}.
\newblock In \emph{Int. Conf. Comput. Vis. (ICCV)}, 2011.

\bibitem[Savinov et~al.(2017)Savinov, Seki, Ladick{\'y}, Sattler, and
  Pollefeys]{Savinov17cvpr}
Nikolay Savinov, Akihito Seki, L'ubor Ladick{\'y}, Torsten Sattler, and Marc
  Pollefeys.
\newblock {Q}uad-networks: Unsupervised learning to rank for interest point
  detection.
\newblock In \emph{{IEEE} Conf. Comput. Vis. Pattern Recog. (CVPR)}, 2017.

\bibitem[Schonberger et~al.(2017)Schonberger, Hardmeier, Sattler, and
  Pollefeys]{Schonberger17cvpr}
Johannes~L. Schonberger, Hans Hardmeier, Torsten Sattler, and Marc Pollefeys.
\newblock Comparative evaluation of hand-crafted and learned local features.
\newblock In \emph{{IEEE} Conf. Comput. Vis. Pattern Recog. (CVPR)}, pages
  1482--1491, 2017.

\bibitem[Shi and Tomasi(1994)]{Shi94cvpr}
Jianbo Shi and Carlo Tomasi.
\newblock Good features to track.
\newblock In \emph{{IEEE} Conf. Comput. Vis. Pattern Recog. (CVPR)}, pages
  593--600, 1994.
\newblock \doi{10.1109/CVPR.1994.323794}.

\bibitem[Simo-Serra et~al.(2015)Simo-Serra, Trulls, Ferraz, Kokkinos, Fua, and
  Moreno-Noguer]{Simoserra15iccv}
Edgar Simo-Serra, Eduard Trulls, Luis Ferraz, Iasonas Kokkinos, Pascal Fua, and
  Francesc Moreno-Noguer.
\newblock Discriminative learning of deep convolutional feature point
  descriptors.
\newblock In \emph{Int. Conf. Comput. Vis. (ICCV)}, 2015.

\bibitem[Sivic and Zisserman(2003)]{Sivic03iccv}
J.~Sivic and A.~Zisserman.
\newblock {Video Google}: a text retrieval approach to object matching in
  videos.
\newblock In \emph{Int. Conf. Comput. Vis. (ICCV)}, 2003.
\newblock \doi{10.1109/ICCV.2003.1238663}.

\bibitem[Tardioli et~al.(2015)Tardioli, Montijano, and Mosteo]{Tardioli15iros}
Danilo Tardioli, Eduardo Montijano, and Alejandro~R. Mosteo.
\newblock Visual data association in narrow-bandwidth networks.
\newblock In \emph{IEEE/RSJ Int. Conf. Intell. Robot. Syst. (IROS)}, pages
  2572--2577, 2015.
\newblock \doi{10.1109/IROS.2015.7353727}.

\bibitem[Wei et~al.(2018)Wei, Zhang, Gong, and Zheng]{Wei18cvpr}
Xing Wei, Yue Zhang, Yihong Gong, and Nanning Zheng.
\newblock Kernelized subspace pooling for deep local descriptors.
\newblock In \emph{{IEEE} Conf. Comput. Vis. Pattern Recog. (CVPR)}, 2018.

\bibitem[Yi et~al.(2016)Yi, Trulls, Lepetit, and Fua]{Yi16eccv}
Kwang~Moo Yi, Eduard Trulls, Vincent Lepetit, and Pascal Fua.
\newblock {LIFT}: Learned invariant feature transform.
\newblock In \emph{Eur. Conf. Comput. Vis. (ECCV)}, pages 467--483, 2016.

\bibitem[Zagoruyko and Komodakis(2015)]{Zagoruyko15cvpr}
Sergey Zagoruyko and Nikos Komodakis.
\newblock Learning to compare image patches via convolutional neural networks.
\newblock In \emph{{IEEE} Conf. Comput. Vis. Pattern Recog. (CVPR)}, pages
  4353--4361, 2015.
\newblock \doi{10.1109/CVPR.2015.7299064}.

\bibitem[Zhang and Rusinkiewicz(2018)]{Zhang18cvpr}
Linguang Zhang and Szymon Rusinkiewicz.
\newblock Learning to detect features in texture images.
\newblock In \emph{{IEEE} Conf. Comput. Vis. Pattern Recog. (CVPR)}, 2018.

\end{thebibliography}

\clearpage

\section{Supplementary Material}

\subsection{Additional results}

\begin{figure}
  \centering
  \begin{tabular}{ccc}
  \includegraphics[trim=0 0 5 0,clip,width=.352\linewidth]{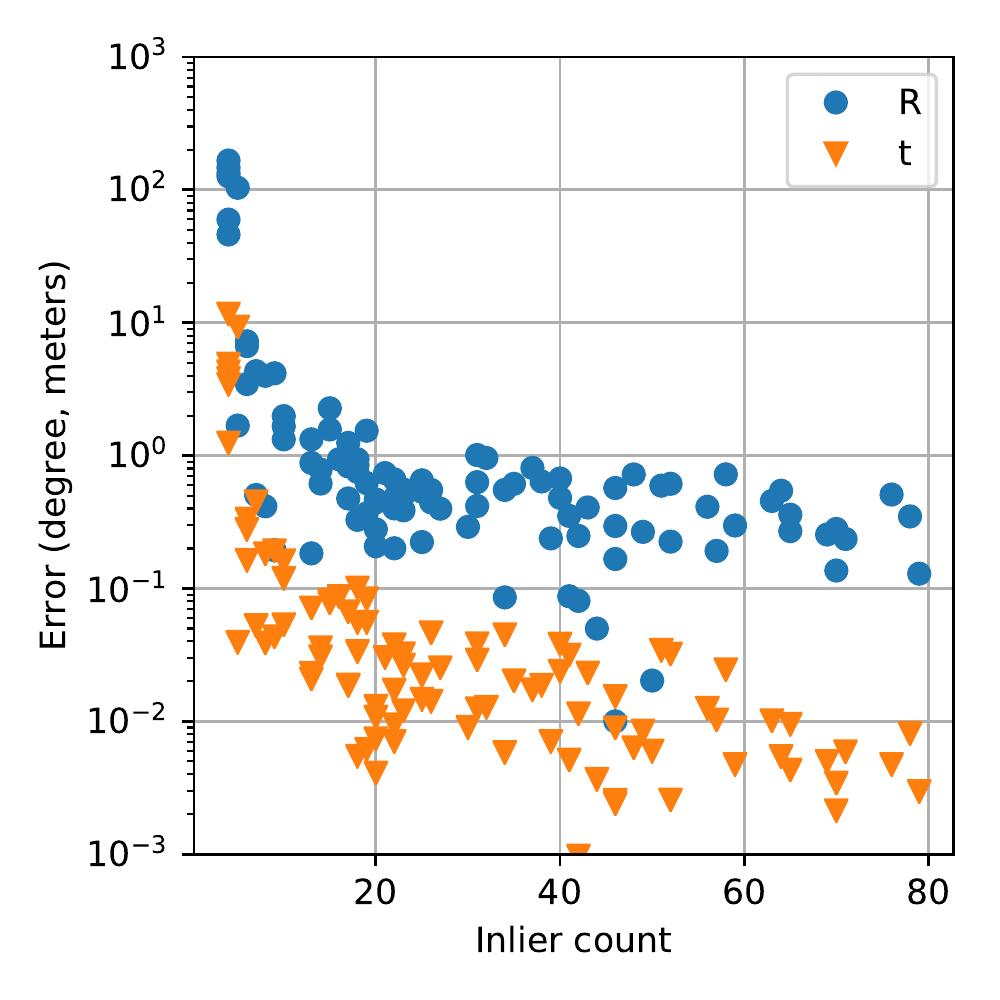}&
  \includegraphics[trim=50 0 5 0,clip,width=.29\linewidth]{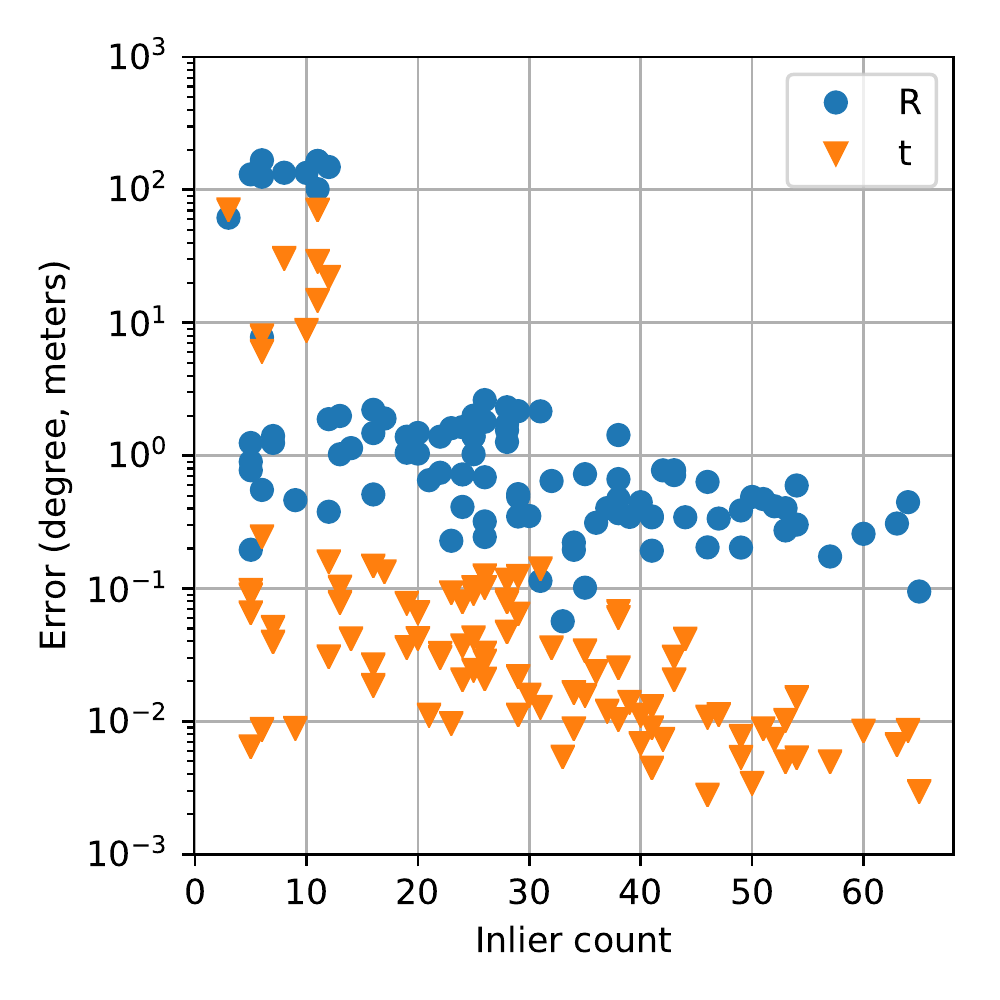}&
  \includegraphics[trim=50 0 5 0,clip,width=.29\linewidth]{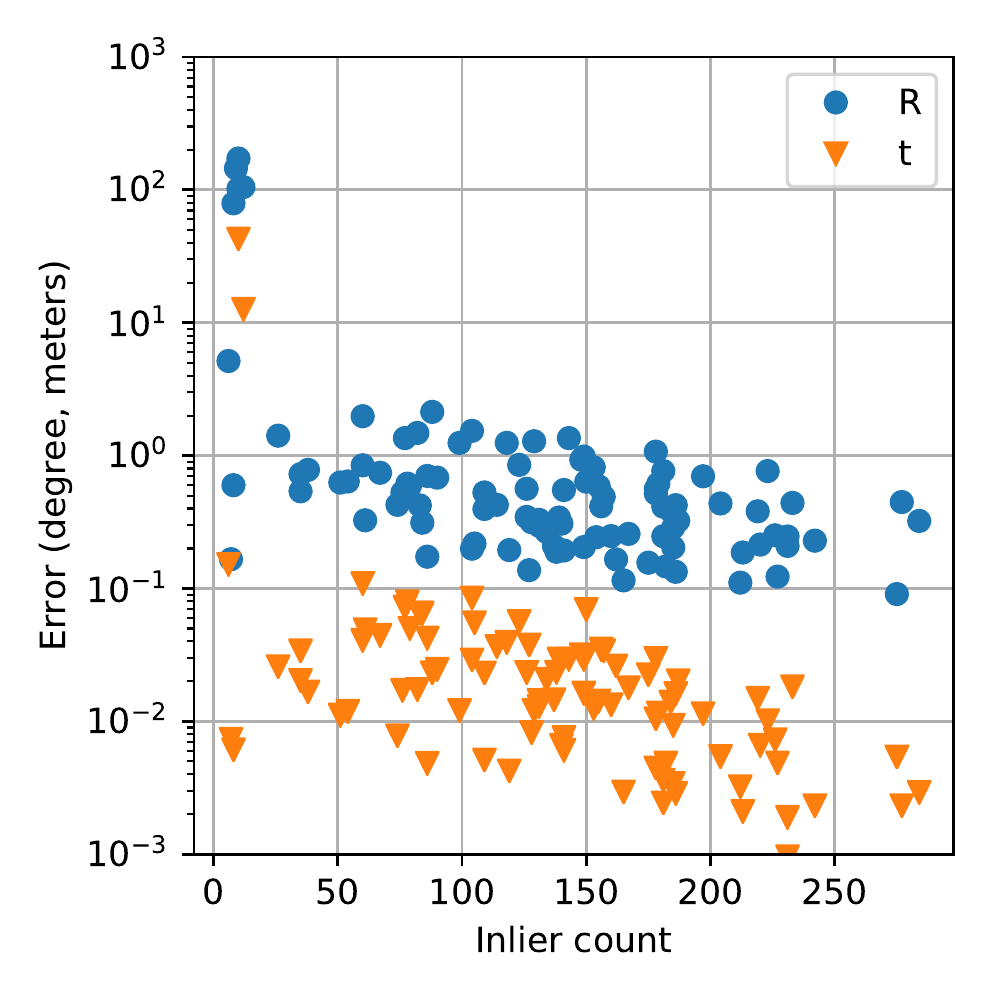}\\
  (a) ours & (b) SIFT, 128 points & (c) SIFT, 500 points
  \end{tabular}
  \caption{Pose error versus inlier count on EuRoC.
  R: rotation error, t: translation error.}
  \label{fig:rt_eu}
\end{figure}

\reffig{fig:rt_eu} shows the equivalent of \reffig{fig:rt} for the EuRoC dataset.
The result is similar except that the rotation error is larger and the translation error smaller.
The latter is likely due to the smaller scene depth of the indoor EuRoC dataset versus the outdoor KITTI dataset.

\begin{figure}
  \centering
  \begin{tabular}{cc}
  \includegraphics[width=.48\linewidth]{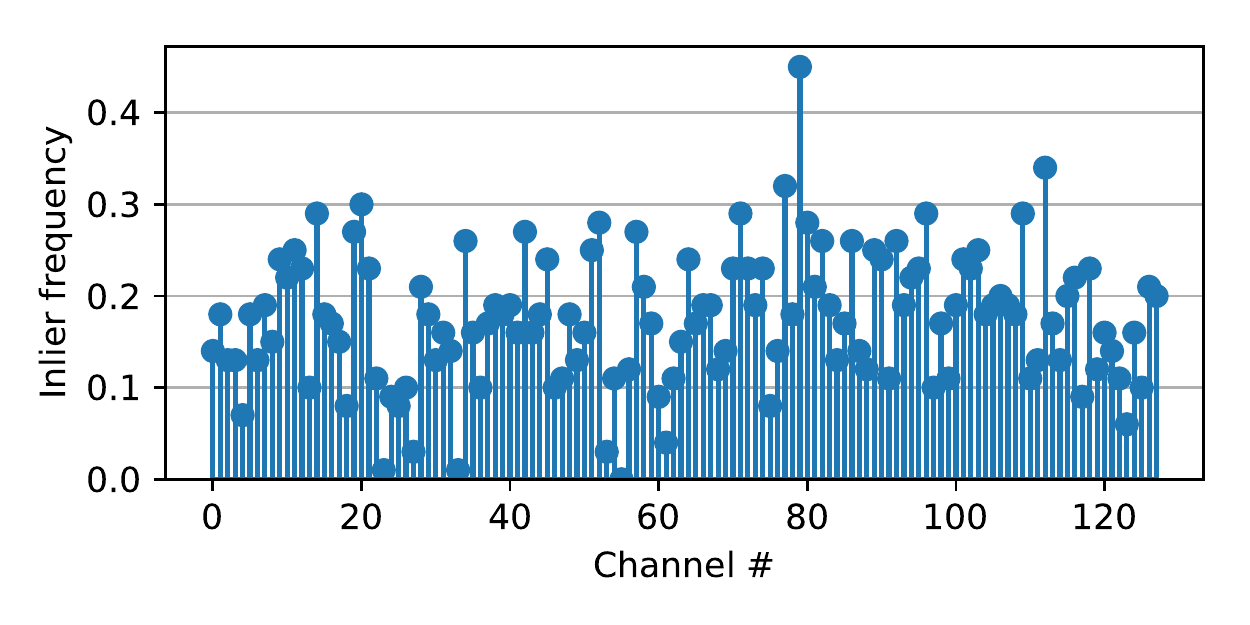} &
  \includegraphics[width=.48\linewidth]{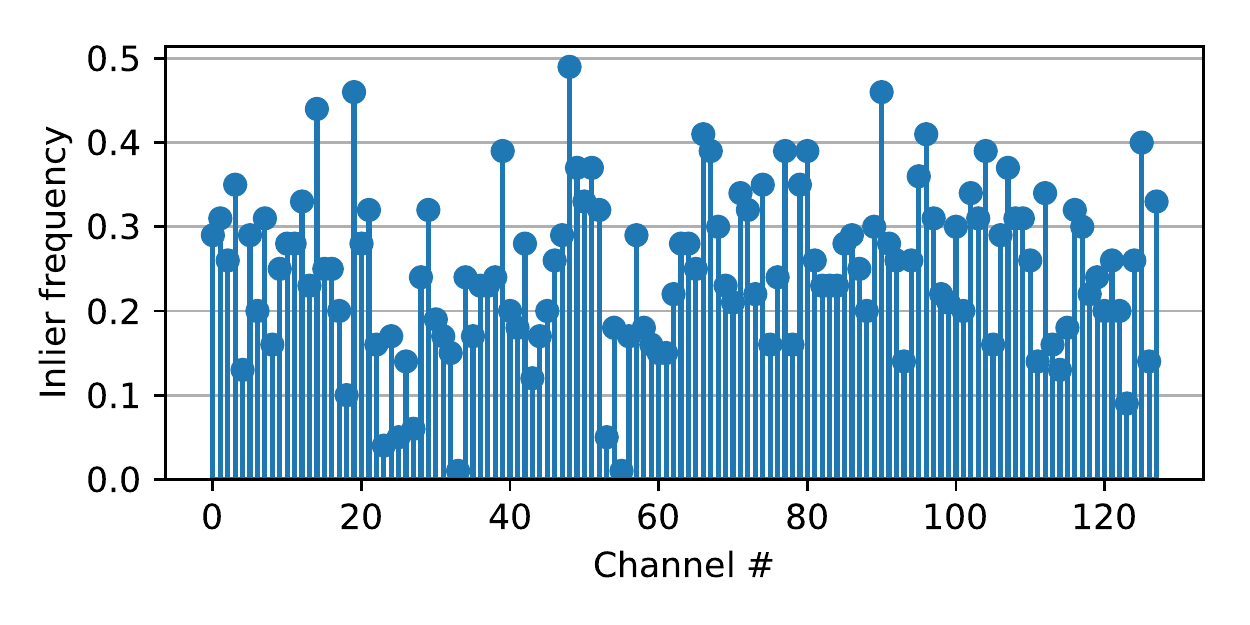} \\
  KITTI & EuRoC \\
  \end{tabular}
  \caption{Distribution of inliers among channels in the testing datasets.}
  \label{fig:inliers}
  \vspace{-5mm}
\end{figure}

\reffig{fig:inliers} shows how inliers are distributed among channels.
As can be seen, the empiric frequency of resulting in an inlier on testing data is well-distributed among the channels, meaning that the training manages to evenly distribute the feature space of good interest points among the channels.

\subsection{CSV files}

A detailed CSV file is attached for each of the two robotic testing dataset.
They contain the columns:
\begin{itemize}
  \setlength\itemsep{-.5em}
  \item {\tt name}: sequence name and the indices of the two images that form the pair.
  \item {\tt dR, dt}: ground truth difference in rotation and translation between the two image frames.
  \item {\tt matching score}: The matching score of our method for the given pair.
  \item {\tt eR, et}: rotation and translation \emph{error} of the relative pose estimate.
\end{itemize}
Both {\tt dR} and {\tt eR} are measured with the geodesic distance, in degrees, which corresponds to the angle of the angle-axis representation of the relative rotation.

\subsection{True relative pose in testing sets}

\begin{figure}
  \centering
  \includegraphics[width=.48\columnwidth]{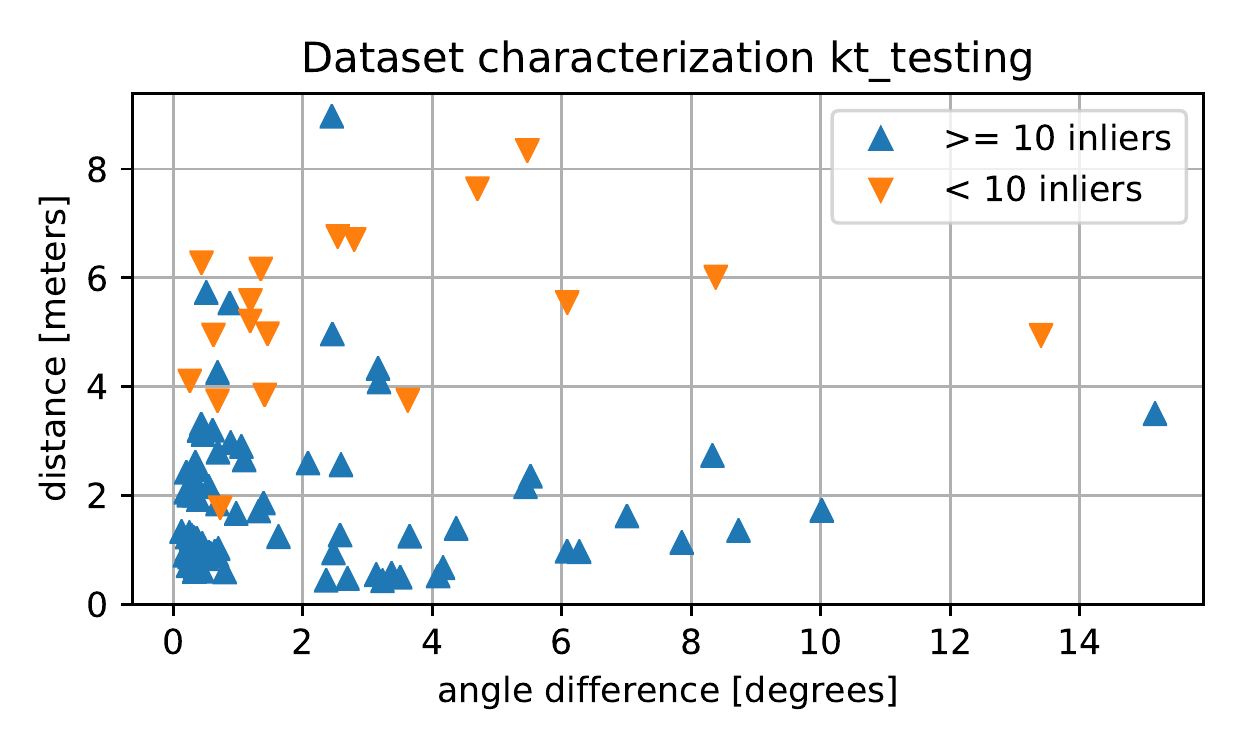}
  \includegraphics[width=.48\columnwidth]{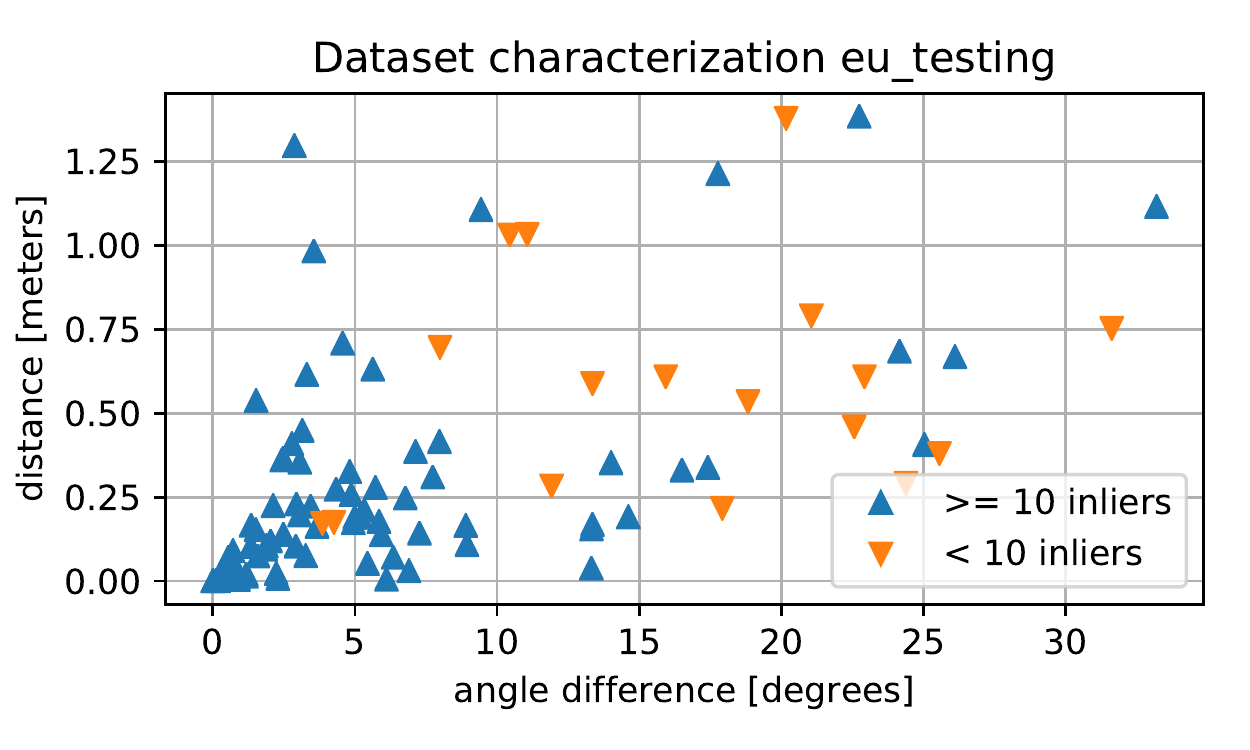}
  \caption{True pose difference plots for KITTI (left) and EuRoC (right).
  Markers indicate whether or not more than $10$ P3P RANSAC inliers have been achieved for a given pair.
  As shown in the paper, this is a good inlier threshold to accept or reject relative pose estimates.} 
  \label{fig:dr_dt}
\end{figure}

To provide an intuition for the robotic testing sets, we plot the distribution of true relative poses in \reffig{fig:dr_dt}
Note the difference between the two datasets:
KITTI exhibits larger distances, while EuRoC exhibits larger orientation differences.
Recall that given the first image, the second image is randomly sampled among subsequent images with a scene overlap of at least $50\%$.

Besides showing the distribution, this plot also indicates which pairs result in $10$ or more P3P RANSAC inliers using our method.
As shown in the paper, these pairs result in a good relative pose estimate.
While the pairs that fail to obtain a good relative pose estimate are typically the ones with larger pose differences, there is no clear boundary between pairs that succeed and pairs that fail.

\subsection{Accuracy versus representation size figures}

In \reffig{fig:moar} (c), (d) we show results for two compression schemes applied to SIFT, principal component analysis (PCA) projection and product quantization \cite{Jegou11pami, Lynen15rss}.
For each method, several lines are plotted for several parameter choices.
For PCA projection, these correspond to projections to $\{3, 5, 10\}$ dimensions, while for product quantization, these correspond to $(m, k) \in \{(1, 256), (2, 16), (4, 4)\}$ (one byte per feature), and $(m, k) \in \{(2, 256), (4, 16), (8, 4)\}$ (two bytes per feature), where $m$ is the amount of quantizations/dimension subdivisions and $k$ the amount of clusters in each quantization.
Both compression methods are trained on the same dataset as our method, TUM mono sequences {\tt 01}, {\tt 02}, {\tt 03} (indoors) and {\tt 48}, {\tt 49}, {\tt 50} (outdoors).

Furthermore, these figures contain results for our network with $64$ and $256$ dimensions.
The $64$ network has the same intermediate channel counts as the standard $128$ network described in \refsec{sec:meth}.
The $256$ network, instead, has all intermediate channel counts doubled.

\subsection{Sequence video}

We visualize the stability of our interest points as well as the correlation of that stability with predicted inlierness probability with a video composed of the score maps extracted over the testing sequences.
Due to size constraints, only $180$ frames of KITTI and $250$ frames of EuRoC are shown.
Note that the interest points are independently extracted in each frame.
Still, several and in particular the high-score interest points behave as if they were directly tracked.
This corresponds to points that would result in correct matches, and gives an idea of what the matching score means.
One observation that can be made in our video is that the response is not always very well-localized.
Future work to address this issue could be to extend the training losses with a loss that favours peakedness \cite{Zhang18cvpr, Ono18nips}.

\end{document}